# Gated Ensemble of Spatio-temporal Mixture of Experts for Multi-task Learning in Ride-hailing System

**Md. Hishamur Rahman**[*]
Department of Civil and Environmental Engineering
University of Utah
Salt Lake City, Utah, United States
Email: hisham.rahman@utah.edu

**Shakil Mohammad Rifaat**
Department of Civil Engineering
Islamic University of Technology
Boardbazar, Gazipur, Bangladesh
Email: srifaat@iut-dhaka.edu

**Soumik Nafis Sadeek**
Department of Transdisciplinary Science and Engineering
Tokyo Institute of Technology
Tokyo, Japan
Email: sadeek.s.aa@m.titech.ac.jp

**Masnun Abrar**
Department of Civil Engineering
Islamic University of Technology
Boardbazar, Gazipur, Bangladesh
Email: masnunabrar@iut-dhaka.edu

**Dongjie Wang**
Department of Computer Science
University of Central Florida
Orange County, Florida, United States
Email: wangdongjie@Knights.ucf.edu

*Corresponding author

# Gated Ensemble of Spatio-temporal Mixture of Experts for Multi-task Learning in Ride-hailing System

**Abstract**

Ride-hailing system requires efficient management of dynamic demand and supply to ensure optimal service delivery, pricing strategies, and operational efficiency. Designing spatio-temporal forecasting models separately in a task-wise and city-wise manner to forecast demand and supply-demand gap in a ride-hailing system poses a burden for the expanding transportation network companies. Therefore, a multi-task learning architecture is proposed in this study by developing gated ensemble of spatio-temporal mixture of experts network (GESME-Net) with convolutional recurrent neural network (CRNN), convolutional neural network (CNN), and recurrent neural network (RNN) for simultaneously forecasting these spatio-temporal tasks in a city as well as across different cities. Furthermore, a task adaptation layer is integrated with the architecture for learning joint representation in multi-task learning and revealing the contribution of the input features utilized in prediction. The proposed architecture is tested with data from Didi Chuxing for: (i) simultaneously forecasting demand and supply-demand gap in Beijing, and (ii) simultaneously forecasting demand across Chengdu and Xian. In both scenarios, models from our proposed architecture outperformed the single-task and multi-task deep learning benchmarks and ensemble-based machine learning algorithms.

## 1. Introduction

In recent years, deep learning has been successfully deployed in various transportation applications, such as ride-hailing system. Ride-hailing system operates based on a mobile application-based platform, which enables it to match the demand-supply dynamically. With the advancement of mobile technologies, ride-hailing services are replacing the conventional taxi services and reshaping the mode choice behavior of passengers. In the last few years, several ride-hailing companies such as Uber, Lyft, DiDi have gained increasing popularity in many cities around the world. To ensure high quality service with real-time information and timely mobility of the passengers, the ride-hailing companies need to forecast different operational metrics (e.g., demand) at the same time in a city as well as across the cities they are operating. In this context, a "task" refers to a specific type of spatio-temporal forecasting of an operational metric. Ride-hailing companies primarily focus on spatio-temporal forecasting of two tasks: demand and supply-demand gap. Addressing these forecasting tasks individually require separate design, maintenance, and updating of forecasting architectures, which is a burden for the ride-hailing companies. Therefore, the question arises—can we model multiple spatio-temporal forecasting tasks in a ride-hailing system with a unified architecture to ease that burden?

Spatio-temporal forecasting has been widely studied in the last decade to enhance the efficiency of ride-hailing companies by incorporating cutting-edge computational tools. Most of these previous studies have focused on spatio-temporal forecasting of demand and supply-demand gap. The essential difference between demand forecasting and supply-demand gap forecasting in ride-hailing services lies in their focus and operational implications. Demand forecasting is primarily concerned with predicting the volume of ride requests in various spatial units and time frames. This task is essential for strategic planning, enabling ride-hailing companies to optimize fleet distribution, schedule drivers efficiently, and anticipate areas of high demand to ensure that customer needs are met promptly. On the other hand, supply-demand gap forecasting focuses on identifying mismatches between the available drivers (supply) and the ride requests (demand). This insight is useful for operational decisions, guiding dynamic pricing mechanisms and driver incentivization programs to mitigate any potential service delays or shortages. While demand forecasting aims at understanding where and when rides are needed, supply-demand gap forecasting focuses on balancing the service's operational efficiency by aligning the supply of rides with passenger demand. However, there exists challenges in both forecasting tasks due to variations in spatio-temporal dependencies (Chiang et al., 2015; Wang et al., 2017), which are found to have converse implications.

Interestingly, spatio-temporal dependencies for forecasting demand and supply-demand gap in a city are modeled using the same type of features, yet separate architectures are utilized for their prediction without considering a common representation from the features. Furthermore, the usual approach in these forecasting architectures is to utilize information that are related to the city of interest only. The same spatio-temporal forecasting task (e.g., demand) in different cities are modeled using the corresponding dataset of the city without considering the correlation among the features of the cities that can provide better inductive bias. Therefore, a multi-task learning framework that has the ability to capture these correlations in a city as well as across the cities with a joint representation can substantially improve the predictive accuracy for spatio-temporal forecasting tasks in a ride-hailing system. Such unified model allows these platforms to make real-time adjustments to their operations by simultaneously analyzing both the current demand and the supply-demand gap. This integrated perspective is useful for rapid decision-making, such as deploying vehicles to areas with anticipated high demand or managing pricing strategies to balance demand and supply. Furthermore, demand forecasting and supply-demand gap analysis share underlying spatio-temporal dependencies and can benefit from common sets of features, such as traffic conditions, weather, time of day, etc. A unified model can exploit these synergies more effectively than separate models, leading to improved forecasting accuracy with a consolidated feature set.

Employing a unified model to simultaneously forecast demand and the supply-demand gap in ride-hailing services offers several advantages over the conventional approach of customizing separate models for each task. Firstly, a unified model facilitates the efficient use of computational resources and data by sharing underlying structures and parameters across tasks, which can lead to significant reductions in both the complexity and the cost of model maintenance and updates. This integrated approach also enables the model to leverage cross-task insights, where the interdependencies between demand and supply-demand gaps can inform and enhance the accuracy of forecasts for both. Additionally, by avoiding the redundancy of processing similar data sets separately for each task, a unified model can more effectively capture and analyze the complex spatio-temporal dynamics that influence ride-hailing demand and supply.

The most popular methods to detect spatio-temporal dependencies are the convolutional neural network (CNN) (LeCun et al., 1999) and the recurrent neural network (RNN) (Williams and Zipser, 1989), which are deep learning techniques with outstanding success in tasks related to computer vision and natural language processing. Earlier studies that applied deep learning for spatio-temporal forecasting generally divide the whole study area into several zones, and the historical data corresponding to the zones are utilized as inputs in the CNN and RNN of the spatio-temporal forecasting architectures. However, for simultaneously forecasting multiple spatio-temporal tasks in a ride-hailing system along with capturing spatio-temporal dependencies in a city as well as across cities, these methods require modification in their forecasting architectures to incorporate multi-task learning, which is not yet done till now.

Inspired by the success of deep learning methods for modeling spatio-temporal forecasting problems, an overlooked aspect is explored in this study—modeling multiple spatio-temporal forecasting tasks of a city or across cities in a ride-hailing system by a multi-task learning architecture. While such multi-task learning architecture have been used in natural language processing (Collobert and Weston, 2008), machine translation (Johnson et al., 2017), speech recognition (Seltzer and Droppo, 2013), computer vision problems (Zhang et al., 2014), and content recommendation (Ma et al., 2018), but rarely been applied in spatio-temporal forecasting problems in ride-hailing system. Previously, for spatio-temporal forecasting problems in ride-hailing system, deep learning was applied to deal only with the problem at hand, which limits the efficiency of deep learning since repeating efforts are required for each problem (Kaiser et al., 2017). In this study, a spatio-temporal multi-task learning architecture with mixture-of-experts is developed in this study for forecasting multiple spatio-temporal tasks in a city as well as across cities. The major contribution of this study are as follows:

(1) We develop a deep multi-task learning architecture for simultaneously forecasting multiple spatio-temporal forecasting tasks in a city as well as across cities by developing gated ensemble of mixture of experts containing convolutional recurrent neural network (CRNN), convolutional neural network (CNN), and recurrent neural network (RNN).
(2) A task adaptation layer is integrated with the multi-task learning architecture, which determines weighting that assists in learning a joint representation for different tasks and aids in interpreting the spatio-temporal multi-task learning models by indicating the contribution of the input features for prediction.
(3) We tested our proposed multi-task learning architecture with real world datasets of Didi for two scenarios of multi-task learning in ride-hailing system: (i) simultaneous forecasting of demand and supply-demand gap in Beijing, and (ii) simultaneous forecasting of demand for Chengdu and Xian. The comparison of the model performance against the benchmark deep learning and machine learning models demonstrated the superiority of our multi-task learning model.

## 2. Literature Review

The Literature review is organized into three sections. First, a discussion of the previous studies on forecasting demand in the taxi and ride-hailing system is provided. Second, a review of the previous studies on forecasting supply-demand gap in ride-hailing system is presented. Finally, a review of the studies that applied multi-task learning in taxi/ride-hailing system is discussed.

### *2.1. Demand Forecasting*

Numerous research papers have been published regarding ride-hailing demand forecasting. Appropriate methodological deployment is one of the most important and critical issues in purpose of forecasting. Li et al. (2012) developed a short-term demand forecasting model based on Wavelet and Support Vector Machine (SVM) which showed powerful predictability and captured non-stationary characteristics of the dynamic ride-sourcing system. Apart from it, Saadi et al. (2017) developed machine learning approach to characterize and forecast ride-hailing travel demand considering traffic, pricing and weather condition. In their case, they found that boosted decision trees had outperformed the prediction accuracy of artificial neural network, random forest, bagged and single decision trees. Ke et al. (2017), for the first time, proposed a deep learning approach to address spatial, temporal and exogenous dependencies of on demand ride-sourcing service simultaneously. They developed a fusion-based convolutional long short-term memory (LSTM) architecture that could capture spatio-temporal correlations of various explanatory factors. Chen et al. (2021) developed UBERNET, a modified convolutional neural network that utilizes dilated causal convolution layers to effectively predict short-term ride-hailing demand. Their proposed model integrates advanced features such as gated activations and skip connections, and employs causal convolutions for handling temporal data and dilated convolutions to manage long-term dependencies. Chen et al. (2022) addressed the challenge of predicting imbalanced ride-hailing demand across varying locations and times using a new bagging method with hexagonal convolutional LSTM, significantly improving accuracy by diversifying demand thresholds and model training strategies.

Graph Convolutional Networks (GCN) are found to be an efficient tool to capture region-based relationships as well as prediction problems by incorporating spatial dependencies. Bai et al. (2019) developed a Cascade Graph CNN to capture both spatial and temporal correlations of passenger demand within a city. Additionally, it incorporated encoder-decoder module to fuse Graph CNN and LSTM to predict the future passenger demand. Similar kind of research was also conducted by Zhou et al. (2020), however, their model did not consider Graph model, instead, they used deep neural network with attention mechanism and kernel density estimation to predict ride-sourcing demand in the pre-predicted city region and time period. Moreover, Wang, et al. (2019) used OD matrix to predict ride-sourcing passenger demand using Graph Convolutions. Also, Xu and Li (2019) proposed graph and time-series learning model to capture ride-sourcing demand where the former learns spatial dependencies and the later captures temporal correlations. While most of the ride-hailing demand forecasting models generally utilized region-based situation awareness or station-based graph representation to capture spatial ride-hailing dynamics in a city, Jin et al. (2020) utilized both simultaneously by combining GCN, variational encoders and sequence to sequence learning to learn and predict spatio-temporal ride-hailing dynamics. In another study, Jin et al. (2022) proposed the Deep Multi-View Spatio-temporal Virtual Graph Neural Network, a novel approach to urban ride-hailing demand prediction that addresses data sparsity and limited long-term dependency learning through an innovative virtual graph model highlighting key demand areas and integrating 1D CNNs, Multi-Graph Attention Networks, and Transformers to enhance spatial and temporal prediction accuracy. Wu et al. (2023) developed another Multi-View Deep Spatio-Temporal Network for improving urban online ride-hailing demand prediction. This advanced model builds upon residual network (ResNet) to analyze temporal dynamics of orders, uses GCN to understand spatial relationships, and incorporates Point of Interest (POI) statistics, weather, and air quality data. It uniquely combines these elements through a multi-view architecture that includes LSTM with an attention mechanism, effectively capturing complex spatio-temporal patterns and offering enhanced forecasting accuracy, as validated by experiments on Didi and Wuhan taxi datasets. However, none of these abovementioned studies considered joint prediction of demand across multiple cities, which can increase the generalization capability of the forecasting models.

### *2.2. Supply-demand Gap Forecasting*

While numerous studies focused on only taxi/ride-hailing demand forecasting, very few studies have attempted to explore ride-hailing supply-demand gap mechanism and forecasting. For instance, Zhang et al. (2017) and Wang (2017) utilized different ensemble techniques to combine decision tree models for forecasting supply-demand gaps in sparse-data situations. Wang et al. (2017) predicted supply-demand gap using deep residual neural network for a given time and geographical area in China. Ling et al. (2019) proposed an ensemble method by combining SVM and ANN using GPS data to predict supply-demand forecasting. Along with predictive power, this two-stage model is able to capture the mobility pattern too.

Moreover, Li & Wang. (2017) proposed LSTM to forecast supply-demand gap with an incorporation of weather and traffic information as well as point of interest (POI). Ke et al. (2017b) developed various hexagon-based CNN to predict supply-demand gap considering spatio-temporal characteristics of ride-hailing services. Furthermore, Said and Erradi (2020) proposed an approach for the purpose of supply-demand gap forecasting not only depending on the raw data but also developing new predictors from the raw gap data using multi-view topological data analysis. Zhang et al. (2020) used fuzzy features in combination with deep learning and attention mechanism to capture ride-hailing supply-demand differences both spatially and temporally. Recently, Song et al. (2022) combined GCN and LSTM architectures for forecasting short-term supply-demand gaps in online car-hailing services. The model also integrated environmental and operational variables such as traffic congestion, weather, air quality, and temperature to enhance prediction accuracy. However, none of these studies jointly forecasted demand and supply-demand gap with a unified architecture, which can provide better inductive bias for predicting both demand and supply-demand gap.

### *2.3. Multi-task Learning for Spatio-temporal Forecasting*

In the domain of ride-hailing systems, accurate prediction of various operational metrics is essential for enhancing service efficiency. In our study, a "task" refers to a specific type of spatio-temporal forecasting of an operational metric within the context of ride-hailing system. For accurate prediction of demand, various researchers attempted to develop multi-task learning architecture using deep learning. Some of these studies utilized multi-task learning to tackle correlation among different regions both spatially and temporally and to capture the influence of confounding factors (e.g., weather). For example, Bai et al. (2019) proposed an end to end multi-task deep learning with historical data by utilizing CNN to capture spatial correlations with high prediction accuracy of passenger demand. Zhang et al. (2019) proposed another multi-task learning method with dynamic time warping algorithm that could capture temporal passenger demand in a multi-zone level in China. Luo et al. (2020) proposed multi-task deep learning framework that outperformed the predictive power of single task deep learning, LSTM, SVM and k-nearest neighbors. This study was unique from other studies as the proposed model could capture causality of demand in various traffic zones along with its multi-zonal predictability power. Kim et al. (2020) developed a step-wise model by combining econometrics models with deep learning where the former could be able to interpret demand and the later could increase the predictive power.

A few studies have developed multi-task learning framework for joint prediction of multiple tasks. For example, Kuang et al. (2019) developed an attention-based LSTM to fuse spatio-temporal historical data with 3D ResNet to predict taxi demand for pick-up and dropping-off, simultaneously. Zhang et al. (2022) also proposed a similar kind of study, however, they proposed multi-task learning with 3-parallel LSTM layers that could co-predict pick-up and dropping-off ride-hailing demand at a time. Ke et al. (2021) proposed a novel deep multi-task multi-graph learning approach to improve the joint prediction of demands for multiple ride-hailing service modes. Their model incorporated multiple multi-graph convolutional (MGC) networks to forecast demand for each service mode and used multi-task learning modules for cross-network knowledge sharing. In another study, Feng et al. (2022) developed a Multi-Task Matrix Factorized Graph Neural Network (MT-MF-GCN) to simultaneously predict zone-based and origin-destination based ride-hailing demand. The MT-MF-GCN integrates a GCN that uses a mixture-model to capture spatial correlations among zones, and a matrix factorization module designed for multi-task demand predictions.

Recently, Zhao et al. (2024) proposed a Temporal and Spatial Intertwined Network (TSIN), a multi-task learning model aimed at improving joint prediction of taxi and ridesourcing demands. TSIN features twin components for capturing intra-mode spatial and temporal dependencies separately for each travel mode and an intertwined component for inter-mode feature exchange. Each component utilizes spatio-temporal blocks, comprised of temporal and spatial convolutional layers, to analyze patterns from historical data at various scales. However, all of these previous studies utilized multi-task learning in their architecture to deal with tasks in a city, while our proposed deep multi-task learning architecture deals with simultaneous forecasting of multiple spatio-temporal forecasting tasks in a city as well as across cities, leveraging useful information from the considered spatio-temporal forecasting tasks to develop a shared representation for better generalizations across all considered spatio-temporal forecasting tasks.

## 3. Preliminaries

This section presents a description of the variables utilized in this study and the problem of spatio-temporal multi-task learning in a ride-hailing system. Spatio-temporal forecasting in ride-hailing system involve time-series data, therefore, historical values of the variables that evolve with time are important indicators for predicting targets. Additionally, POI, time of day, and day of week also have effects on the spatio-temporal forecasting.

*Space-time partitioning*

For aggregation of the variables, the study area is divided into $N$ non-overlapping uniformly sized zones $Z = \{1,2,3,\dots,N\}$ and total time is divided into $T$ time-slots $I = \{1,2,3,\dots,T\}$ of $s$ minutes interval. The rest of the variables are explained based on this space-time partitioning.

*Spatio-temporal variables*

Spatio-temporal variables vary simultaneously in the spatial and temporal dimension. The following types of the spatio-temporal variables are utilized in this paper:

(1) Demand: The ride-hailing demand at all zones during the time-slot $t \in I$ is represented by the vector $\boldsymbol{D}_t \in \mathbb{R}^N$, where $D_{t,z}$ is the total number of successful ride-hailing requests emerging from zone $z \in Z$, and $D_{t,z} \in [0, +\infty)$.
(2) Original demand: The total number of ride-hailing requests from a zone in a time interval is referred to as the original demand, which includes both successfully matched and unanswered ride-hailing requests. The original demand of all zones during the time-slot $t$ is placed in the vector $\boldsymbol{OD}_t \in \mathbb{R}^N$. The demand at zone $z$ during the time-slot $t$ is denoted as $OD_{t,z}$, where $OD_{t,z} \geq D_{t,z}$.
(3) Supply-demand gap: The total number of unanswered ride-hailing requests from a spatial zone in a time-slot is termed as the supply-demand gap. The supply-demand gap of all zones during the time-slot $t$ is expressed as the vector $\boldsymbol{G}_t \in \mathbb{R}^N$. The supply-demand gap of a zone $z$ during the time-slot $t$ is denoted by $G_{t,z}$, where $0 \leq G_{t,z} \leq OD_{t,z}$.

(4) Traffic congestion: The traffic congestion of all zones is denoted by the vector $\boldsymbol{TC}_t \in \mathbb{R}^N$, where $TC_{t,z}$ represents the total number of congested roads belonging to a zone $z$ during the time-slot $t$.

(5) Speed: The average speed of the floating ride-hailing cars of a spatial zone in a time-slot is termed as the speed in that spatial zone. The speed at all zones during the time-slot $t$ is denoted by the vector $\boldsymbol{S}_t \in \mathbb{R}^N$. The speed at a zone $z$ during the time-slot $t$ is denoted by $S_{t,z}$.

*(6)* Accessibility: The accessibility of a spatial zone in a time-slot is measured by the number of ride-hailing cars with passengers crossing that spatial zone. The accessibility of all zones in the time-slot $t$ is expressed by the vector $\boldsymbol{M}_t \in \mathbb{R}^N$. The accessibility of a zone $z$ during the time-slot $t$ is denoted by $M_{t,z}$.

*Weather variables*

The weather variables include the meteorological features that vary randomly across time, but not space. For maintaining consistency of input dimensions in our proposed architectures, these variables may have to be repeated across the zones by utilizing the repeating function $f_{RZ}(\cdot\,;N):\mathbb{R}^{1\times M} \rightarrow \mathbb{R}^{N\times M}$, where $M$ represents the number of feature categories. The following types of weather variables are included in this paper:

(1) Categorical weather variables: For the time-slot $t$, the weather conditions are the categorical weather variables, denoted by the row vector $\boldsymbol{wc}_t \in \mathbb{R}^{1\times C}$ consisting of $C$ weather categories (e.g., sunny, rainy, snowy, etc.), where each weather category is encoded by one-hot encoding, i.e., $wc_{c,t} \in \{0,1\}$.

(2) Continuous weather variables: For the time-slot $t$, a continuous weather variable is expressed by the vector $\boldsymbol{wt}_t \in \mathbb{R}$. The continuous weather variables utilized in this study are temperature, particulate matter, dew point, humidity, cloud cover, wind speed, and visibility.

*Context variables*

The context variables are either periodic or fixed across time. For spatio-temporal forecasting, these variables are either repeated across the zones by applying the repeating function $f_{RZ}$ or repeated across the time-slots by utilizing the repeating function $f_{RT}(\cdot\,;T):\mathbb{R}^N \rightarrow \mathbb{R}^{N\times 1\times T}$. These are as follows:

(1) Temporal context: Following (Ke et al., 2017), exploratory data analysis of the trends in demand and supply-demand gap revealed two types of periodic contexts: time-of-day and day-of-peak. A time-slot $t$ of a day belongs to one of the three 8-hour time-of-day intervals: sleep (first 8-hour), peak (mid-8-hour), and off-peak (last 8-hour), denoted by the row vector $\boldsymbol{cd}_t \in \mathbb{R}^{1\times 3}$, where each interval $i$ is one-hot encoded, i.e., $cd_{i,t} \in \{0,1\}$. Furthermore, a time-slot $t$ falls into one of the day-of-week categories: weekday or weekend, represented by the vector $\boldsymbol{cw}_t \in \mathbb{R}$, where $cw_t \in \{0,1\}$. The temporal context vectors $\boldsymbol{cd}_t$ and $\boldsymbol{cw}_t$ are repeated across the zones to form the time-of-day matrix $\boldsymbol{CD}_t \in \mathbb{R}^{N\times3}$ and the day-of-week vector $\boldsymbol{CD}_t \in \mathbb{R}^{N}$ respectively.

*(2)* Spatial context: The spatial context refers to the number of POIs across the zones, denoted by the vector $\boldsymbol{cp}_z \in \mathbb{R}^{N}$, which is fixed across time. Therefore, it is repeated across the time-slots with the repeating function $f_{RT}$ to form the spatial context vector for the time-slot $t$, represented by the vector $\boldsymbol{CP}_t \in \mathbb{R}^{N}$.

*Problem Statement*

Our spatio-temporal forecasting problem can be formulated as follows:

For $n$ spatio-temporal tasks in a ride-hailing system, given, the historical data of spatio-temporal variables and weather variables up to $b$-th previous time-slot starting from $t-1$, and known data of context variables at the time-slot $t$, it is required to predict the ground truth vector $\boldsymbol{A}_t$ at the time-slot $t$ for $n$ spatio-temporal forecasting tasks simultaneously.

## 4. Methodology

This section first provides a brief description of task adaptation layer and mixture of experts for dealing with multi-task learning in our proposed architecture. Furthermore, developments of gated mixture of experts for the different components, i.e., CNN, RNN, and Conv-RNN are also explained. Finally, our proposed GESME-Net architecture is explained. We introduced several innovative methods in our proposed architecture for improving spatio-temporal forecasting while multi-task learning in ride-hailing system. First, we employ a task-adaptive feature weighting mechanism that dynamically adjusts the importance of features based on their relevance to each task, accommodating features of any dimension. Second, we have developed a gated convolutional mixture of experts (Conv-ME) that leverages one-dimensional convolutions to detect spatial dependencies in a multi-task learning setting. Additionally, to capture temporal dependencies, we innovated a gated recurrent mixture of experts (GRU-ME) using GRU, enhancing the model's ability to handle temporal dependencies for multi-task learning. Finally, we modified the cell structure of traditional recurrent neural networks (RNNs) to incorporate one-dimensional convolution operations and integrated the ConvRNN within a gated mixture of experts framework, which leverages multiple ConvRNN units for detecting spatio-temporal dependencies while multi-task learning in ride-hailing system.

### *4.1. Task Adaptation Layer*

Multi-task learning methods involve capturing complex interaction among features through multiple intermediate layers between inputs and outputs in such a way that a joint representation for all task is learnt. However, less important input features may hinder the learning process in multi-task learning. To address this issue, task adaptation layer based on a one-to-one linear layer (Borisov et al., 2019; Li et al., 2016; Lu et al., 2018) is utilized in our proposed architecture for learning joint feature importance in multi-task learning. This paper employs a task-adaptive feature weighting, i.e., adaptable with feature of any dimension, is utilized for predicting set of $n$ tasks $P = \{1,2,3, \dots, n\}$, which can be expressed by the following function in Eq. (1):

$$f_{Weighting}\left(\boldsymbol{X}_t^p\right) = \boldsymbol{X}_t^p \odot \sigma(\boldsymbol{W}^{(FI)}) \tag{1}$$

The operator $\odot$ in Eq. (1) indicates Hadamard product, i.e., elementwise multiplication, between any input $\boldsymbol{X}_t^p$ for the time-slot $t$ of the task $p \in P$ and the outputs of the activation function $\sigma$ (e.g., linear, sigmoid, rectified linear unit (ReLU), hyperbolic tangent, etc.) for the weights $\boldsymbol{W}^{(FI)}$. For providing a fair advantage to all the features, i.e., all features are considered as equally important, training of the task adaptation layer is initialized with uniform weights $\boldsymbol{W}^{(FI)} \sim U(-\gamma, +\gamma)$, where $\gamma$ will depend on the type of activation function utilized in Eq. (1).

The theoretical justification of the task adaptation layer utilized in this paper is similar to (Borisov et al., 2019). The weights are updated during the training process, assigning larger weights to more important features and smaller weights to less important features. To ensure that the task adaptation layer correctly captures the contribution of the input features, the weights and biases of the subsequent layers must not become zero during the training process. This is maintained by utilizing the L2-norm of regularization for the weights and biases used after the task adaptation layer in the training algorithm. Furthermore, to ensure sparsity of $\boldsymbol{W}^{(FI)}$, the L1-norm of regularization for $\boldsymbol{W}^{(FI)}$ is included in the training algorithm.

### *4.2. Mixture of Experts*

Mixture of experts were initially developed as an ensemble method (Jacobs et al., 1991) and later utilized as stacked layers in deep learning models (Eigen et al., 2014). Instead of single neural network, a mixture of experts contains subnetworks called experts and a gating network that learns the probabilities of the experts. A mixture of experts (ME) model can be expressed by the following function:

$$f_{ME}(\boldsymbol{X}_t) = \sum_{i=1}^{m} f_{gate}(\boldsymbol{X}_t)_i \times f_{\text{expert}}^{i}(\boldsymbol{X}_t) \tag{2}$$

where the functions $f_{gate}$ and $f_{\text{expert}}$ represents the gating and expert network respectively, $m$ represents the number of expert networks utilized, and $\sum_{i=1}^{m} f_{gate}(\boldsymbol{X}_t)_i = 1$.

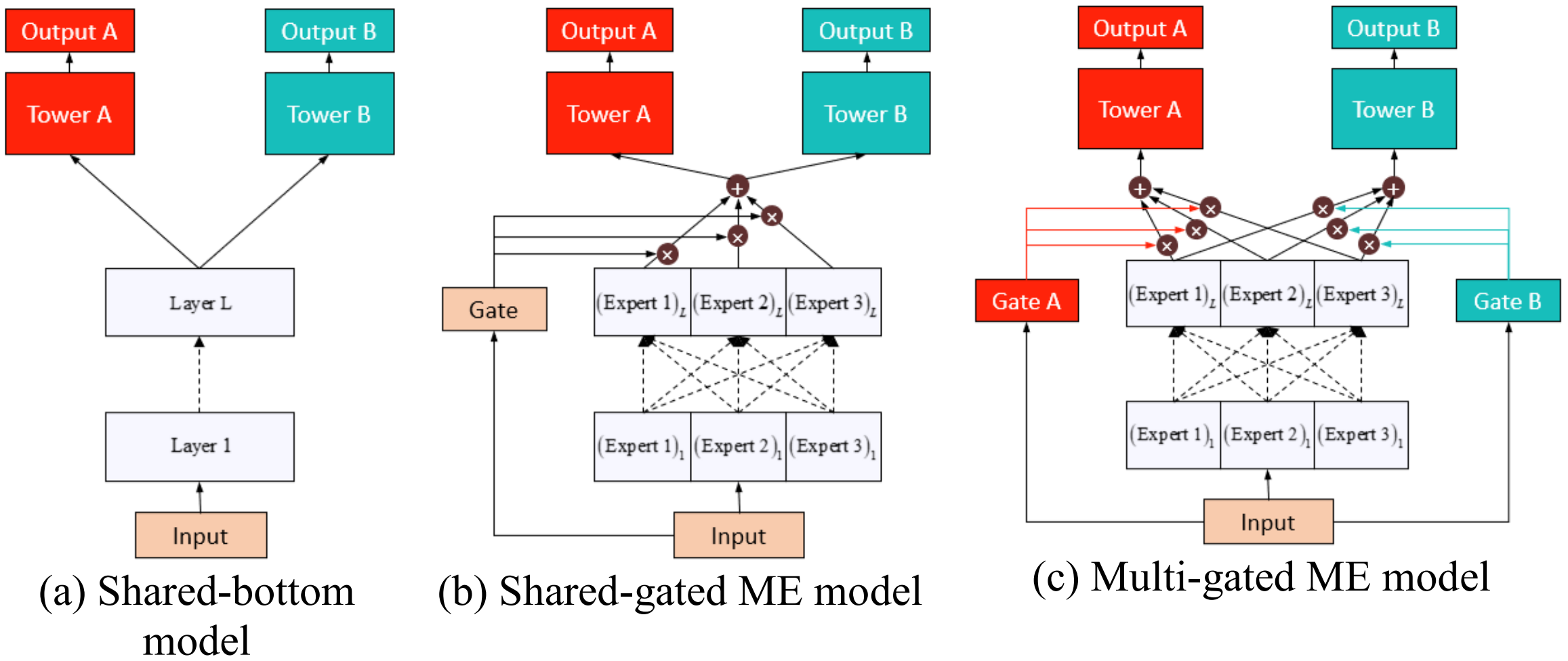

(a) Shared-bottom model (b) Shared-gated ME model (c) Multi-gated ME model

Fig. 1. Multi-task learning models

The concept of multi-task learning through gated mixture of experts (Ma et al., 2018) is developed based on the commonly used shared-bottom neural network (SBNN) (Caruana, 1997) for multi-task learning, which is shown in **Fig 1 (a)**. In the SBNN, the bottom layers utilize shared parameters for all tasks and then task-specific layers (i.e., tower) are introduced following the shared layers. The mathematical formulation of a shared-bottom multi-task learning model can be expressed as follows:

$$\boldsymbol{Y}_t^p = h^p(f_{shared}(\boldsymbol{X}_t)) \tag{3}$$

where, $\boldsymbol{Y}_t^p$ is the represents the output vector of the task $p \in P$ at the time-slot $t$, $f_{shared}$ and $h^p$ represents shared bottom layers and the task-specific layers respectively.

However, SBNN are only found to be useful in multi-task learning settings with similar tasks (Ruder, 2017). Therefore, instead of the shared bottom layers, (Ma et al., 2018) utilized gated mixture of experts, as shown in **Fig. 1 (b)** and **Fig. 1 (c)**. The gating networks produce task-specific softmax probabilities for each expert by processing the input features, providing diversity and flexibility in learning different tasks through the experts. The outputs of the ensembled experts are finally forwarded to the task-specific layers. Therefore, by modifying Eq. (2)-(3), the gated mixture of experts for multi-task learning are expressed through the Eq. (4)-(5).

$$\boldsymbol{Y}_t^p = h^p(f_{ME}^p(\boldsymbol{X}_t)) \tag{4}$$

$$f_{ME}^p(\boldsymbol{X}_t) = \sum_{i=1}^{m} f_{gate}^p(\boldsymbol{X}_t)_i \times f_{\text{expert}}^i(\boldsymbol{X}_t) \tag{5}$$

Although such ensemble of experts is proven to learn task relationships in multi-task learning, the experts utilized are feedforward neural networks that are unable to learn spatio-temporal dependencies. Therefore, we develop gated mixture of experts based on CNN, RNN, and Conv-RNN for learning spatio-temporal dependencies in multi-task learning.

*4.2.1. Gated Convolutional Mixture of Experts*

The CNN is a specialized neural network for detecting spatial dependencies by various types of filters (i.e., weights and bias) sliding over and convolving with the input. However, spatial dependencies in spatio-temporal forecasting not only depends on neighboring zones, but also on distant zones (e.g., functional similarity, transportation connectivity) (Geng et al., 2019). Furthermore, the spatial adjacency information is sometimes anonymized in the ride-hailing datasets due to confidentiality reasons. Therefore, a one-dimensional CNN is utilized in this paper, based on recent findings that showed the applicability of CNN to learn from scrambled images (Brendel and Bethge, 2019). The one-dimensional CNN filters are slid over only across the flattened spatial dimension of the input with spatio-temporal feature columns, as shown in **Fig. 2**. To detect same kind of spatial features with a filter, parameters of a filter are shared across the zones. The output feature vector of a filter $k$ in a one-dimensional CNN layer can be expressed as follows:

$$\boldsymbol{Z}^{(k)} = \sigma\left(\boldsymbol{X}_t * \boldsymbol{W}^{(k)} + \boldsymbol{b}^{(k)}\right) \tag{6}$$

where $*$ is one-dimensional convolution operator, $\boldsymbol{X}_t$ is the input to be convolved with the filter $k$, $\boldsymbol{Z}^{(k)}$ refers to the output feature vector for the filter, $\boldsymbol{W}^{(k)}$ serves as the shared weight matrix of the filter, and $\boldsymbol{b}^{(k)}$ is the shared bias vector of the filter. Therefore, the one-dimensional CNN layer with $K$ filters applied to the input $\boldsymbol{X}_t$ with $N$ zones and $F$ features can be represented by the function $f_{Conv1D}: \mathbb{R}^{N \times F} \rightarrow \mathbb{R}^{N \times K}$.

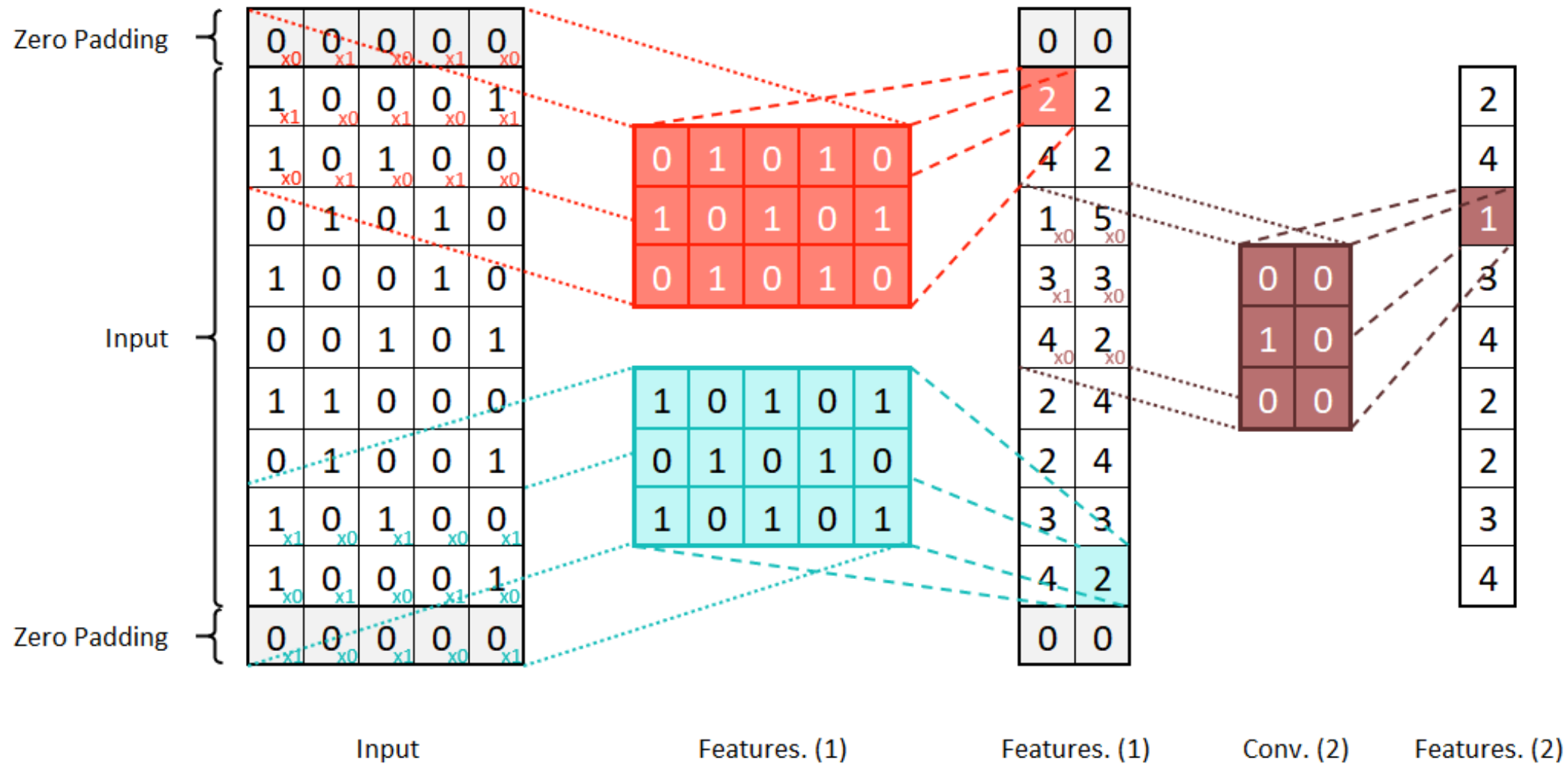

Fig. 2. One-dimensional CNN

We developed gated convolutional mixture of experts for detecting special dependencies in multi-task learning by utilizing one-dimensional convolution instead of feedforward neural networks as the experts in Eq. (5), which is expressed as follows:

$$f^{p}_{Conv-ME}(\boldsymbol{X}_t) = \sum_{i=1}^{m} f^{p}_{gate}(\boldsymbol{X}_t)_i \times f^{i}_{Conv1D}(\boldsymbol{X}_t) \tag{7}$$

*4.2.2. Gated Recurrent Mixture of Experts*

The RNN is an exclusive architecture for detecting temporal dependencies from time-series data where the features of one time-slot are correlated with the features of the previous time-slots. In comparison to the non-recurrent connections in conventional neural networks, the RNN has recurrent connections, i.e., the outputs of the hidden layer neurons from the previous time step of a sequence are utilized with the inputs of the current time step, which can be expressed by Eq. (8):

$$\boldsymbol{h}_t = \sigma(\boldsymbol{U}\boldsymbol{x}_t + \boldsymbol{W}\boldsymbol{h}_{t-1} + \boldsymbol{b}) \tag{8}$$

where $\boldsymbol{x}_t \in \mathbb{R}^F$ is the input vector at the current time step containing $F$ features, $\boldsymbol{h}_t \in \mathbb{R}^H$ and $\boldsymbol{h}_{t-1} \in \mathbb{R}^H$ refer to the output vectors of size $H$ representing hidden layer neurons of current and previous time step respectively, $\boldsymbol{U} \in \mathbb{R}^{H \times F}$ and $\boldsymbol{W} \in \mathbb{R}^{H \times F}$ serve as the weights for the input of the current step and the outputs of the previous step respectively, and $\boldsymbol{b} \in \mathbb{R}^H$ is the bias vector.

However, repeated multiplication of the recurrent hidden layer weight matrix during training results in the vanishing/exploding gradient problem in RNN that limits the storing of long-term information, which can be tackled through LSTM (Hochreiter and Schmidhuber, 1997) and GRU (Cho et al., 2014) by including additive updates in the hidden layer. Although LSTM and GRU are found to be similar in terms of performance (Jozefowicz et al., 2015), training GRU requires lesser time, which is an advantage for learning large number of tasks with multi-task learning. Therefore, GRU is chosen for developing gated recurrent mixture of experts in our study. The GRU cell, as shown in **Fig. 3**, modifies the recurrent structure in Eq. (8) through Eq. (9)-(12):

$$\boldsymbol{z}_t = \sigma\left(\boldsymbol{U}^{(z)}\boldsymbol{x}_t + \boldsymbol{W}^{(z)}\boldsymbol{h}_{t-1} + \boldsymbol{b}^{(z)}\right) \tag{9}$$

$$\boldsymbol{r}_t = \sigma\left(\boldsymbol{U}^{(r)}\boldsymbol{x}_t + \boldsymbol{W}^{(r)}\boldsymbol{h}_{t-1} + \boldsymbol{b}^{(r)}\right) \tag{10}$$

$$\widetilde{\boldsymbol{h}}_t = tanh\left(\boldsymbol{U}^{(h)}\boldsymbol{x}_t + \boldsymbol{r}_t \odot \boldsymbol{W}^{(h)}\boldsymbol{h}_{t-1} + \boldsymbol{b}^{(h)}\right) \tag{11}$$

$$\boldsymbol{h}_t = \boldsymbol{z}_t \odot \widetilde{\boldsymbol{h}}_t + (1 - \boldsymbol{z}_t)\boldsymbol{h}_{t-1} \tag{12}$$

where $\boldsymbol{z}_t$, $\boldsymbol{r}_t$, and $\widetilde{\boldsymbol{h}}_t$ are update gate, reset gate, and new states respectively, containing corresponding input weights, recurrent weights, and bias vector. The recurrent layer with gated recurrent units can be denoted by the function $f_{GRU} : \mathbb{R}^F \rightarrow \mathbb{R}^H$.

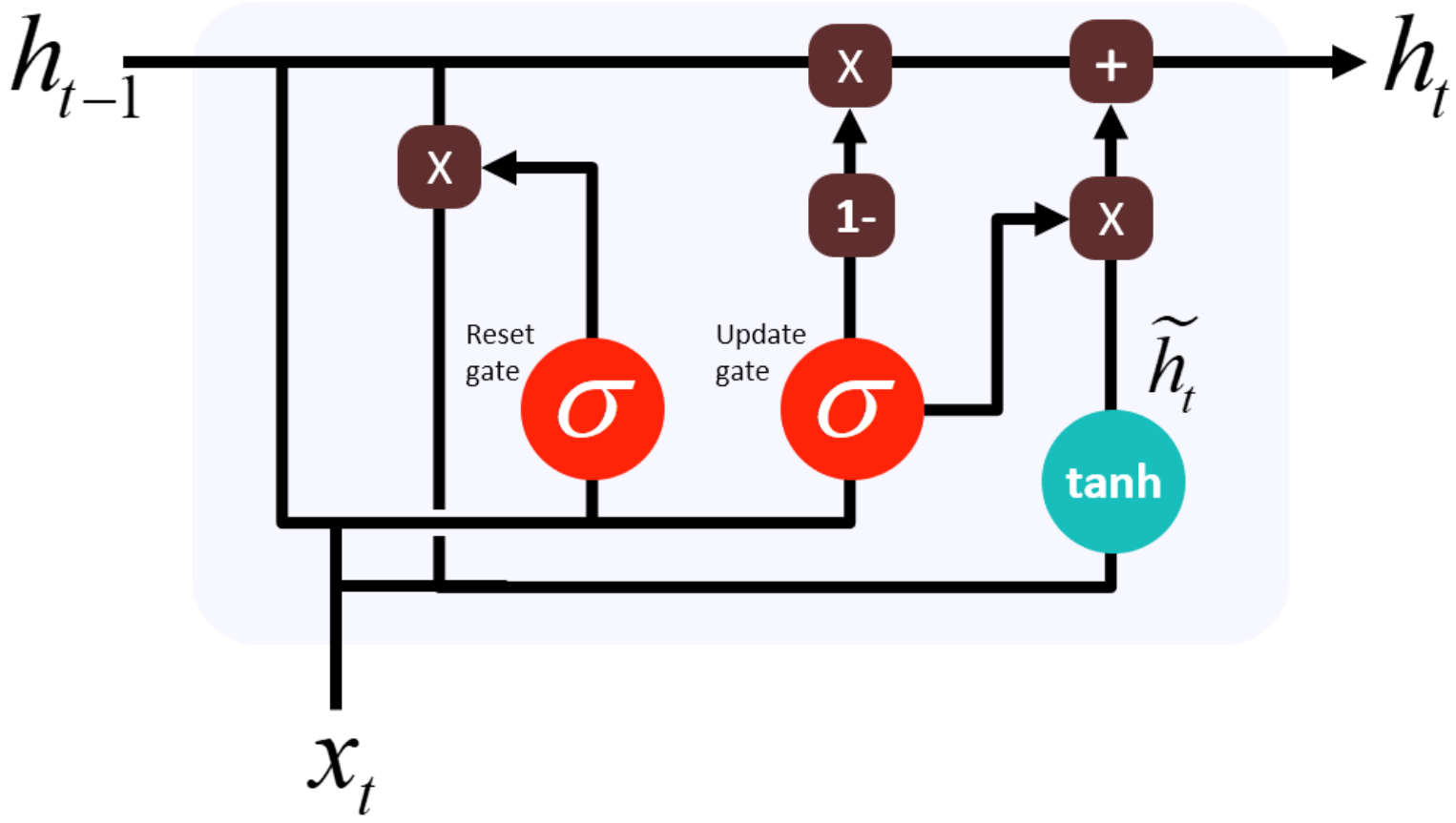

Fig. 3. A GRU cell

For detecting temporal dependencies in spatio-temporal features, instead of a vector, the input of GRU at a time step is a matrix $\boldsymbol{X}_t \in \mathbb{R}^{N \times F}$ including $N$ zones, which requires the same GRU layer to be distributed across the zones independently with shared parameters. Therefore, the GRU layer is applied to the $N$ zones of the input matrix $\boldsymbol{X}_t$ with the function $f_{ZoneDist(GRU)} : f_{GRU}(x_t) \rightarrow f_{GRU}(\boldsymbol{X}_t)$.

The gated recurrent mixture of experts for learning temporal dependencies in multi-task learning is developed by replacing the non-recurrent gates and experts in Eq. (5) with recurrent gates and recurrent experts as follows:

$$f_{GRU-ME}^{p}(\boldsymbol{x}_t) = \sum_{i=1}^{m} f_{GRUgate}^{p}(\boldsymbol{x}_t)_i \times f_{GRU}^{i}(\boldsymbol{x}_t) \tag{13}$$

$$f_{ZoneDist(GRU)-ME}^{p}(\boldsymbol{X}_t) = \sum_{i=1}^{m} f_{ZoneDist(GRU)gate}^{p}(\boldsymbol{X}_t)_i \times f_{ZoneDist(GRU)}^{i}(\boldsymbol{X}_t) \tag{14}$$

where the function $f_{GRU-ME}^{p}$ and $f_{ZoneDist(GRU)-ME}^{p}$ represents gated mixture of experts and zone-distributed gated mixture of experts respectively with $f_{GRUgate}^{p}$ and $f_{ZoneDist(GRU)gate}^{p}$ are the corresponding recurrent gating layers for producing softmax probabilities following a GRU layer.

*4.2.3. Gated Convolutional Recurrent Mixture of Experts*

Convolutional recurrent layers are exclusively utilized to detect spatial and temporal dependencies simultaneously from spatio-temporal features. The incorporation of convolutional recurrent layers is aimed at enhancing the model's ability to detect more complex spatio-temporal patterns that CNN and RNN are unable to capture. By integrating convolution operations into the recurrent structure, convolutional recurrent layers allow for processing of spatial information at each time step, maintaining spatial coherence throughout the temporal sequence analysis. To prevent vanishing/exploding gradient problems in convolutional recurrent layers, convolutional LSTM (Shi et al., 2015) and gated convolutional recurrent unit (Wang et al., 2020) have been developed. However, the use of LSTM/GRU in the convolutional recurrent layer makes it computationally expensive due to large number of output units (Sak et al., 2014) and does not provide better model performance other than that they only ease the training process (Collins et al., 2017). Rather, a convolutional recurrent layer with a vanilla RNN (ConvRNN) can easily tackle the vanishing/exploding gradient problem by utilizing a linear activation function, while being computationally cheaper. A ConvRNN can be formulated by modifying the cell structure of the vanilla RNN in Eq. (8) to take spatio-temporal features as inputs and replacing the matrix multiplication with a convolution operator, as shown in **Fig. 4**. Therefore, a one-dimensional ConvRNN is implemented in this study, which is expressed as follows:

$$\boldsymbol{H}_t = ReLU\left(\boldsymbol{U}^{(c)} * \boldsymbol{X}_t + \boldsymbol{W}^{(c)} * \boldsymbol{H}_{t-1} + \boldsymbol{b}^{(c)}\right) \tag{15}$$

where $\boldsymbol{X}_t \in \mathbb{R}^{N\times F}$ is the input at the current time step containing $N$ zones and $F$ features, $\boldsymbol{H}_t \in \mathbb{R}^{N\times K}$ and $\boldsymbol{H}_{t-1} \in \mathbb{R}^{N\times K}$ refer to the output of current and previous time step respectively, $\boldsymbol{U}^{(c)}$ and $\boldsymbol{W}^{(c)}$ serve as the filters for the input of the current step and the outputs of the previous step respectively, and $\boldsymbol{b}^{(c)} \in \mathbb{R}^{N\times K}$ is the bias. Therefore, a convolutional recurrent layer processing $B$timesteps can be denoted as the function $f_{ConvRNN}: \mathbb{R}^{N\times F\times B} \rightarrow \mathbb{R}^{N\times K\times B}$.

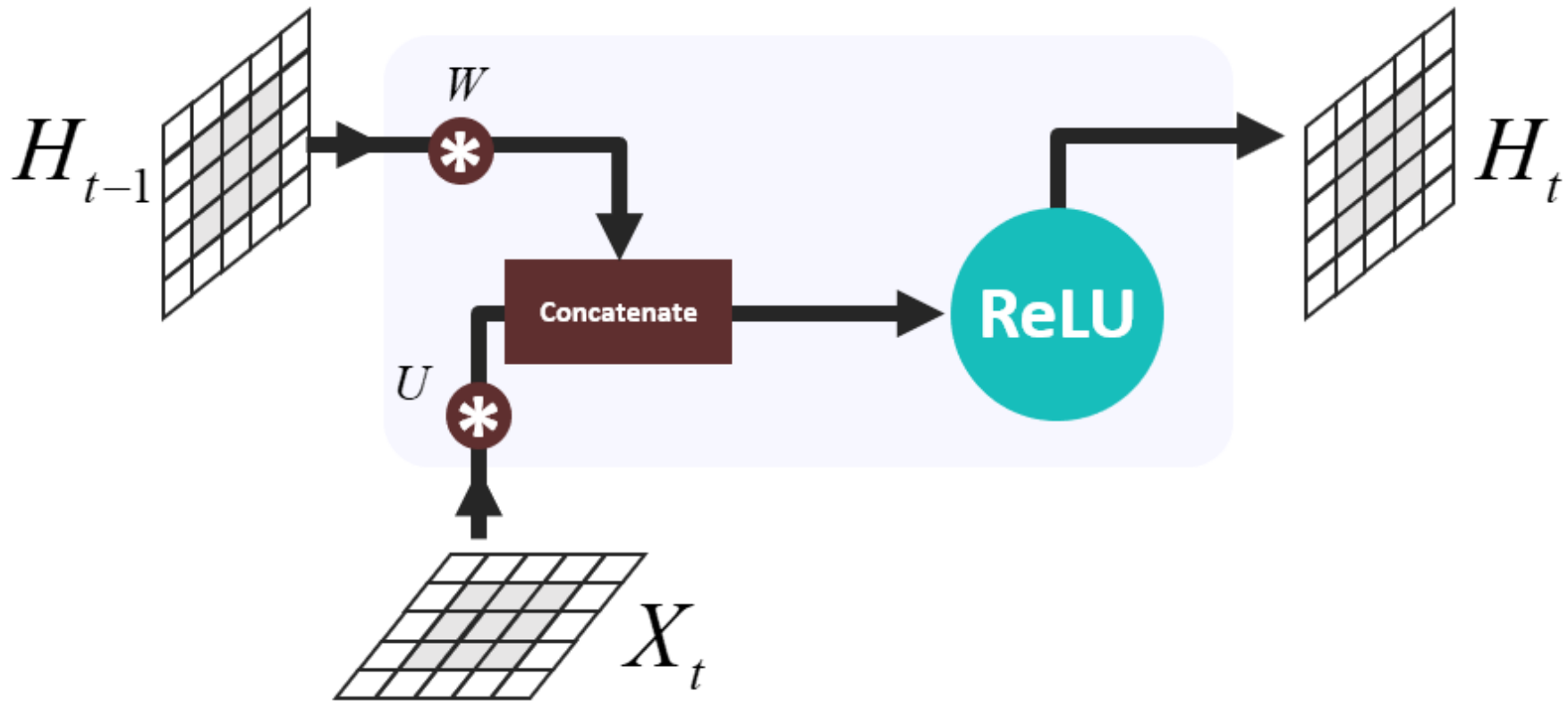

Fig. 4. A ConvRNN cell

The gated convolutional recurrent mixture of experts for multi-task learning is developed by utilizing the ConvRNN as follows:

$$f^{p}_{ConvRNN-ME}(\boldsymbol{X}_t) = \sum_{i=1}^{m} f^{p}_{ConvRNNgate}(\boldsymbol{X}_t)_i \times f^{i}_{ConvRNN}(\boldsymbol{X}_t) \tag{16}$$

where the function $f^{p}_{ConvRNN-ME}$ represents gated mixture of experts and $f^{p}_{ConvRNNgate}$ is the recurrent gating layer for producing softmax probabilities from the outputs of a ConvRNN layer.

### *4.3. Architecture of GESME-Net*

The workflow of the GESME-Net architecture is shown in **Fig. 5**. Initially, all inputs are weighted through the task adaptation layer to adjust their importance based on the unique operational demands of ride-hailing, such as rapid fluctuations in demand and varying impacts of external factors like weather and POI. This layer helps prioritize features dynamically, enhancing the model's responsiveness to real-time changes.

The weighted spatio-temporal variables in the ride-hailing data are then passed through the layers of gated convolutional recurrent mixture of experts to learn task-specific spatial and temporal dependencies simultaneously from the spatio-temporal variables. The hidden layers in the gated convolutional recurrent mixture of experts helps to learn the complex spatio-temporal dependencies from ride-hailing data across cities and across different features. The outputs from these hidden layers are then weighted at their respective gate to specialize on the corresponding ride-hailing forecasting task.

To further specialize the learning process, spatio-temporal features in the ride-hailing data are separately processed through a gated convolutional mixture of experts and a zone-distributed gated recurrent mixture of experts for detecting additional spatial and temporal dependencies that were not detected by the gated convolutional recurrent mixture of experts. This ensures that time-independent spatial correlation among the zones and zone-specific temporal dependencies are learned for spatio-temporal forecasting in ride-hailing system.

The weighted weather features are separately passed through a gated recurrent mixture of experts to learn the task-specific dependencies on the weather features. This ensures that the impact of weather on ride-hailing demand and supply-demand gap are accurately captured. Finally, the task-specific outputs of the different mixture of experts are combined with the weighted temporal context features to learn the effect of weekday/weekend and time-of-day on demand and supply-demand gap through a fully connected layer (i.e., dense) before making the final prediction $\boldsymbol{O}_t$ in the tower layer.

Techniques such as concatenating, reshaping, and permuting dimensions are utilized in the architecture to adapt the input requirement of the architecture. To concatenate different vectors/matrix, the concatenation function $f_{Concatenate}$ is applied (e.g., $f_{Concatenate}(\boldsymbol{P}\,;\boldsymbol{Q}) : \mathbb{R}^{N,N} \rightarrow \mathbb{R}^{N\times 2}$, where $\boldsymbol{P} \in \mathbb{R}^N$ and $\boldsymbol{Q} \in \mathbb{R}^N$ are the vectors to be concatenated). The function $f_{Permute}$ is applied to permute the dimensions of a tensor (e.g., $f_{Permute}(\boldsymbol{V}\,;(2,3,1)) : \mathbb{R}^{C\times D\times E} \rightarrow \mathbb{R}^{D\times E\times C}$, where $\boldsymbol{V} \in \mathbb{R}^{C\times D\times E}$ is the tensor to be permuted with the provided permuted ordering of the dimensions). To meet the requirement of time steps for the zone-distributed recurrent mixture of experts, corresponding inputs are reshaped through the reshaping function $f_{Reshape}(\boldsymbol{J}\,;B) : \mathbb{R}^{N\times A} \rightarrow \mathbb{R}^{N\times B\times\left(\frac{A}{B}\right)}$, where $\boldsymbol{J}$ is the matrix with $A$ features to be reshaped to $B$ time-steps.

According to the problem mentioned in section 3, the prediction target is denoted as the ground truth $\boldsymbol{A}_t$. The training of the GESME-Net involves minimizing the mean squared error between the predicted values and the ground truth, which is achieved through the task-specific loss function $f_{Loss}^{p}$ in Eq. (17). The objective of the overall loss function $f_{Loss}$ applied in our architecture including regularization terms can be expressed as Eq. (18):

$$f_{Loss}^{p}(\boldsymbol{O}_t, \boldsymbol{A}_t) = \|\boldsymbol{O}_t - \boldsymbol{A}_t\|_2^2 \tag{17}$$

$$\min_{W^{(A)},W^{(FI)},b} f_{Loss} = \sum_{p=1}^{n} f_{Loss}^{p} + \alpha\left\|\boldsymbol{W}^{(A)}\right\|_2^2 + \beta\left\|\boldsymbol{W}^{(FI)}\right\|_1 \tag{18}$$

where $\boldsymbol{A}_t \in \mathbb{R}^N$ is the ground truth vector for a task $p$, $\boldsymbol{W}^{(A)}$ refers to all parameters of the GESME-Net except the task adaptation layer parameter $\boldsymbol{W}^{(FI)}$, and $\alpha, \beta$ are the regularization parameters. The L1- and L2-norm of regularization are utilized in accordance with the requirements of the task adaptation layer, which also assists in avoiding overfitting issues.

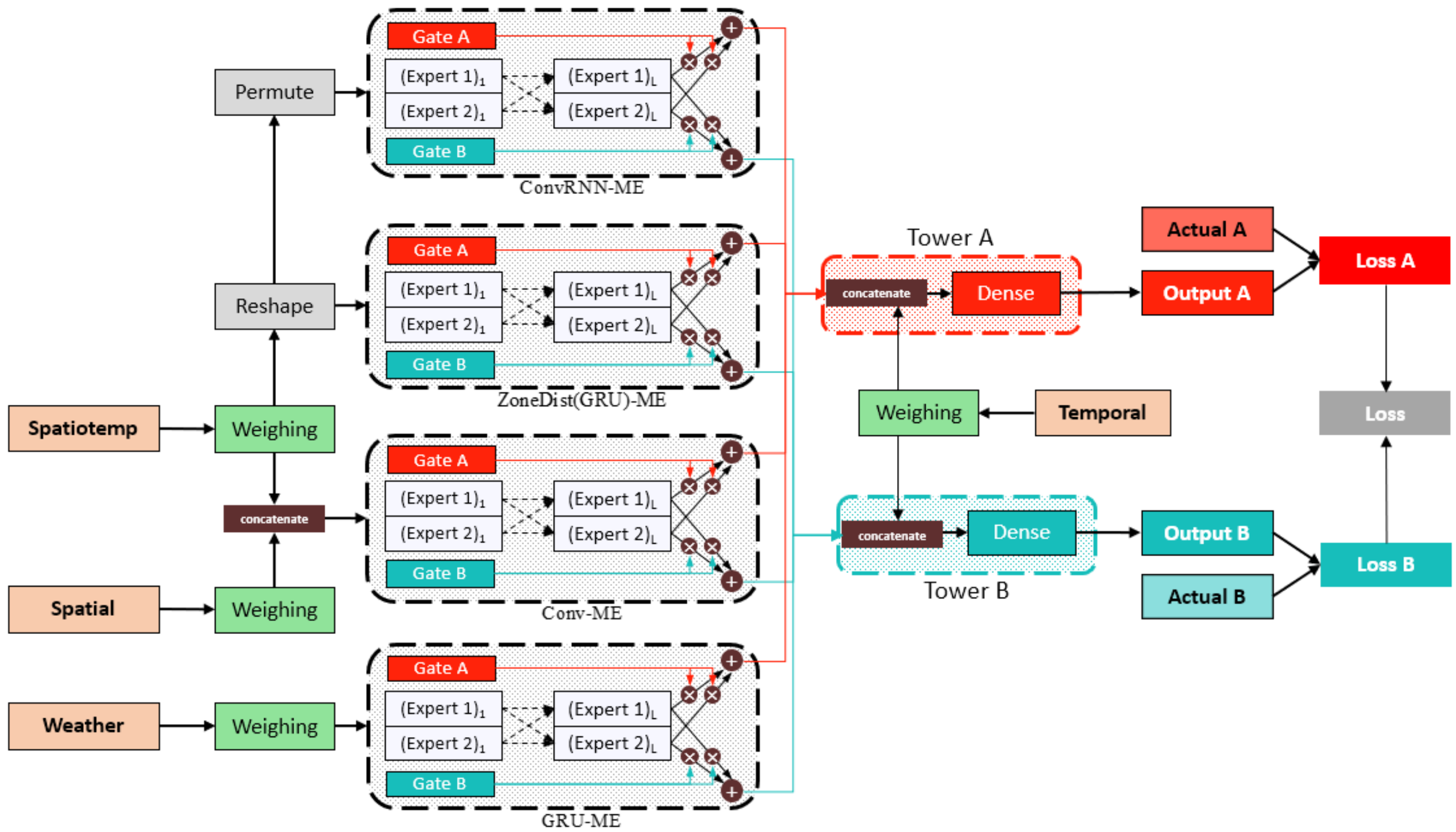

Fig. 5. GESME-Net Architecture

## 5. Experiments and Discussions

This section presents the experiments on real-world data and the discussion of their results. Furthermore, model ablation and model interpretation of the proposed GESME-Net is also provided. Finally, a sensitivity analysis of the GESME-Net hyperparameters are provided at the end of the section.

### *5.1. Data Description and Preprocessing*

In this section, the data description and preprocessing for the two forecasting scenarios are provided. The first scenario involves jointly predicting demand and supply-demand gap with data from Beijing and the second scenario involves jointly predicting demand across cities with data from Chengdu and Xian.

#### *5.1.1. Scenario-1: Forecasting Different Tasks in a City*

Publicly released ride-hailing dataset (Didi, 2016) of Didi Chuxing from Beijing is selected for simultaneously forecasting two spatio-temporal tasks, i.e., demand and supply-demand gap. Didi Chuxing divided each city into several square zones by applying geohashing and identified each zone with a unique ID, anonymizing the adjacency information among the zones.

We utilized the Beijing dataset spanning from 1$^{st}$ January 2016 to 20$^{th}$ January 2016. To construct time-series data for each zone, the total timespan is divided into equal interval time-slots. Considering the small length of the datasets and focusing on short-term forecasting, each day in the datasets is therefore divided into 144 time-slots of 10 minutes interval. Furthermore, the total time-slots in a zone are split into training, validation, and testing sets for our experiments. Around 30% of the time-slots are reserved for validation and testing, and the rest of the data are used in training the models.

The trip dataset contain information around 8.5 million ride-hailing orders. The order information provides anonymized information of each order including driver ID, passenger ID, and, trip origin and destination geohashes. Besides, date and time of the orders are also available in the datasets. The unfulfilled orders are marked by a 'null' driver ID, which is useful information for calculating the supply-demand gaps in a zone for a time-slot. Furthermore, in order to find the demand and quantity supplied of each zone from the order information, the orders including 'null' driver IDs and orders excluding 'null' driver IDs are aggregated, respectively, for a timeslot. For both the datasets, around 50 percent of the time-slots are found to have zero supply-demand gaps, around 17 percent of which are due to zero original demand, indicating that orders in around 33 percent time-slots are fully matched by the platform. The zone-wise distribution of the original demand and the supply-demand gap, is shown in **Fig. 6 (a) and Fig. 6 (b)** respectively. In general, relatively higher original demand zones tend to show a higher supply-demand gap in the dataset.

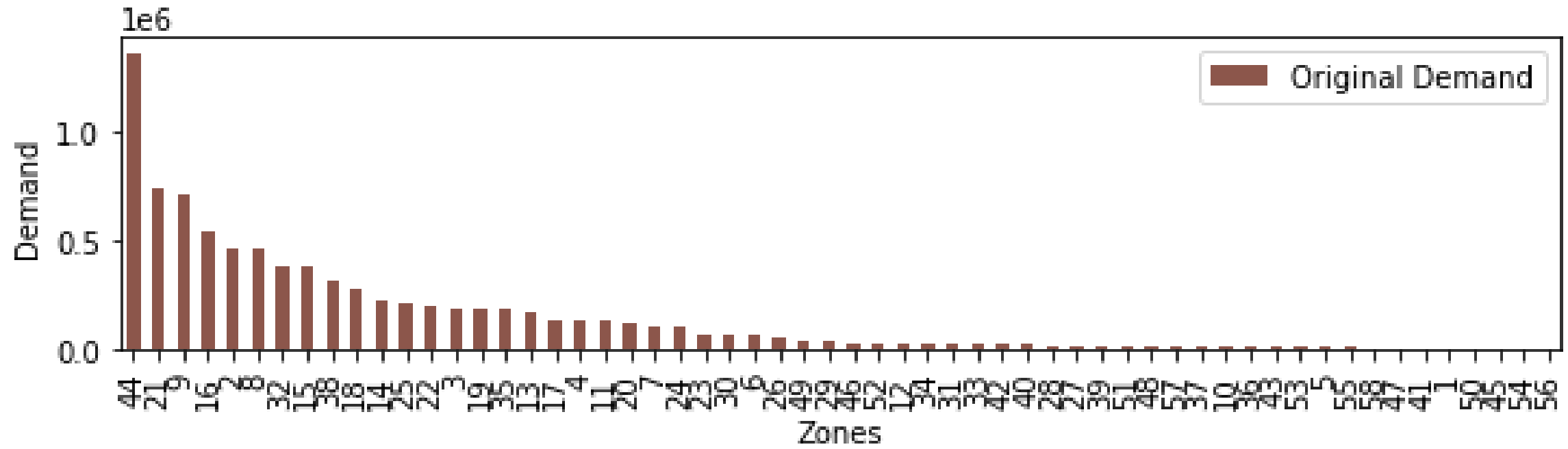

(a) Distribution of original demand across zones

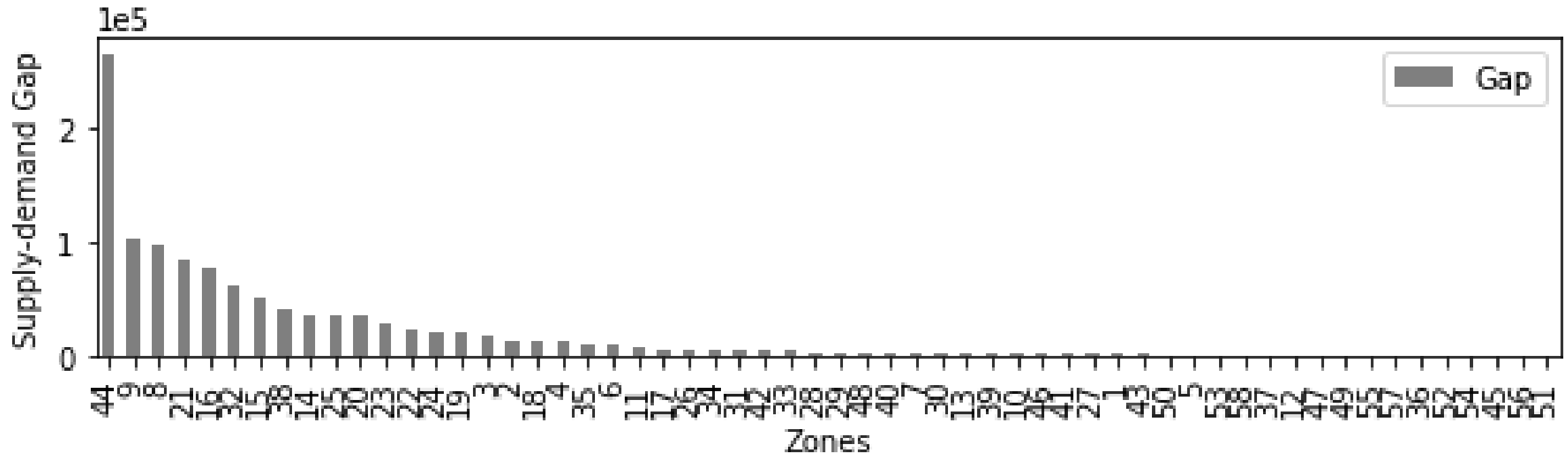

(b) Distribution of supply-demand gap across zones

Fig. 6. Distribution of demand and supply-demand gap in Beijing

City-level weather variables, i.e., weather category, temperature, $PM_{2.5}$ for Beijing are provided with the ride-hailing trip dataset. Furthermore, the zone-wise time-invariant POI information is also provided with the trip dataset. The zone-wise number of facilities for different POI types is aggregated to get the total POI per zone. The descriptive statistics of the input variables in the dataset are summarized in **Table 1**.

**Table 1.** Descriptive statistics of input variables for Beijing (Scenario-1)

| Features | Beijing Dataset | | | |
|---|---|---|---|---|
| | **Mean** | **Std.** | **Min.** | **Max.** |
| Demand | 42.68 | 103.29 | 0.00 | 1.00 |
| Supplied | 35.10 | 74.96 | 0.00 | 1.00 |
| Gap | 7.57 | 45.00 | 0.00 | 0.00 |
| Congestion | 785.98 | 996.20 | 1.00 | 168.00 |
| POI | 296487.3 | 521902.1 | 2988.0 | 50796.0 |
| Temperature | 6.22 | 3.69 | 0.00 | 3.00 |
| $PM_{2.5}$ | 118.63 | 52.08 | 26.00 | 81.00 |
| Weather | Categorical | | 0 | 1 |
| Hourtype | Categorical | | 0 | 1 |
| Daytype | Binary | | 0 | 1 |

### *5.1.2. Scenario-2: Forecasting Same Task Across Different Cities*

Trajectory datasets from Chengdu and Xian under Didi Chuxing GAIA open dataset initiative (Didi, 2018) are selected for simultaneously learning same spatio-temporal task, i.e., demand across different cities. Since the datasets provided raw trajectories only, data processing is used to extract the spatio-temporal variables and external weather and POI datasets are used to extract the required weather and POI variables.

We divided both Chengdu and Xian city into 10×10 square zones as shown in **Fig. 7 (a)** and **Fig. 7 (b)** respectively. Both datasets span from 1$^{st}$ October 2016 to 30$^{th}$ November 2016. Each day in the datasets is divided into 96 time-slots of 15 minutes interval. Furthermore, the total time-slots in a zone are split into training, validation, and testing sets for our experiments. Around 30% of the time-slots are reserved for validation and testing, and the rest of the data are used in training the models.

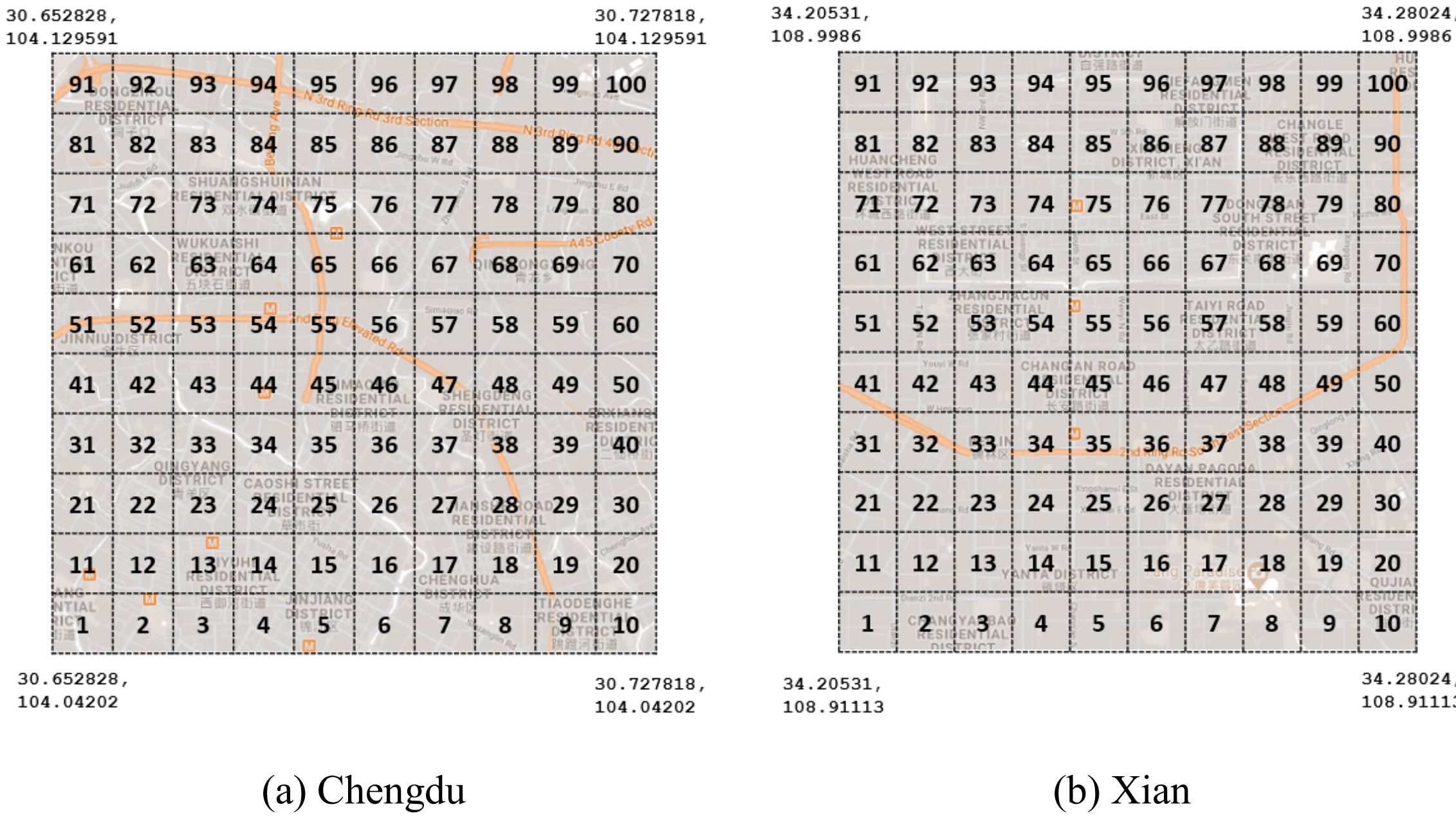

(a) Chengdu (b) Xian

Fig. 7. Spatial partitioning of Chengdu and Xian

The Chengdu and Xian datasets contain anonymized trajectories of around 11.75 and 6.72 million ride-hailing trips respectively. The accessibility of a zone in a time-slot is calculated by counting the trajectories that fall in that zone. The average speed of a ride-hailing vehicle in a zone is found by extracting speed of each ride-hailing vehicle, calculating distance from the trajectory portion that falls in that zone and dividing with the corresponding time to cover that distance. The demand of a zone is calculated by extracting starting point of the trajectories and aggregating them in respective time-slot. For both the datasets, around 12 percent of the time-slots are found to be zero-demand time-slots. The zone-wise distribution of the demand for Chengdu and Xian are shown in **Fig. 8 (a)** and **Fig. 8 (b)** respectively.

City-level weather variables (i.e., weather category, temperature, humidity, visibility, cloud cover, and wind speed) for Chengdu and Xian are collected from Dark Sky (Apple, 2020) at 15 minutes interval. The zone-wise time-invariant POI information for Chengdu and Xian are extracted from Gaode Map (Center, 2017). The zone-wise number of facilities for different POI types is aggregated to get the total POI per zone. The descriptive statistics of the input variables in the dataset are summarized in **Table 2**.

**Table 2.** Descriptive statistics of input variables for Chengdu and Xian (Scenario-2)

| **Features** | **Chengdu** | | | | **Xian** | | | |
|---|---|---|---|---|---|---|---|---|
| | **Mean** | **Std.** | **Min.** | **Max.** | **Mean** | **Std.** | **Min.** | **Max.** |
| Demand | 20.07 | 29.56 | 0.00 | 2.00 | 11.47 | 12.54 | 0.00 | 2.00 |
| Speed | 34.98 | 11.31 | 0.00 | 28.00 | 30.27 | 9.35 | 0.00 | 24.62 |
| Accessibility | 126.57 | 118.48 | 0.00 | 26.00 | 75.80 | 65.21 | 0.00 | 18.00 |
| POI | 1373.31 | 1431.16 | 20.00 | 413.00 | 1373.30 | 846.48 | 154.00 | 914.00 |
| Temperature | 61.20 | 8.40 | 34.73 | 55.35 | 52.08 | 11.43 | 23.10 | 44.62 |
| CloudCover | 0.44 | 0.26 | 0.00 | 0.26 | 0.37 | 0.42 | 0.00 | 0.00 |
| Humidity | 0.82 | 0.14 | 0.33 | 0.73 | 0.79 | 0.18 | 0.22 | 0.67 |
| Visibility | 3.97 | 1.88 | 0.04 | 2.32 | 2.97 | 1.75 | 0.06 | 1.55 |
| WindSpeed | 3.06 | 2.67 | 0.00 | 0.57 | 4.72 | 3.22 | 0.00 | 2.24 |
| Weather | Categorical | | 0 | 1 | Categorical | | 0 | 1 |
| Hourtype | Categorical | | 0 | 1 | Categorical | | 0 | 1 |
| Daytype | Binary | | 0 | 1 | Binary | | 0 | 1 |

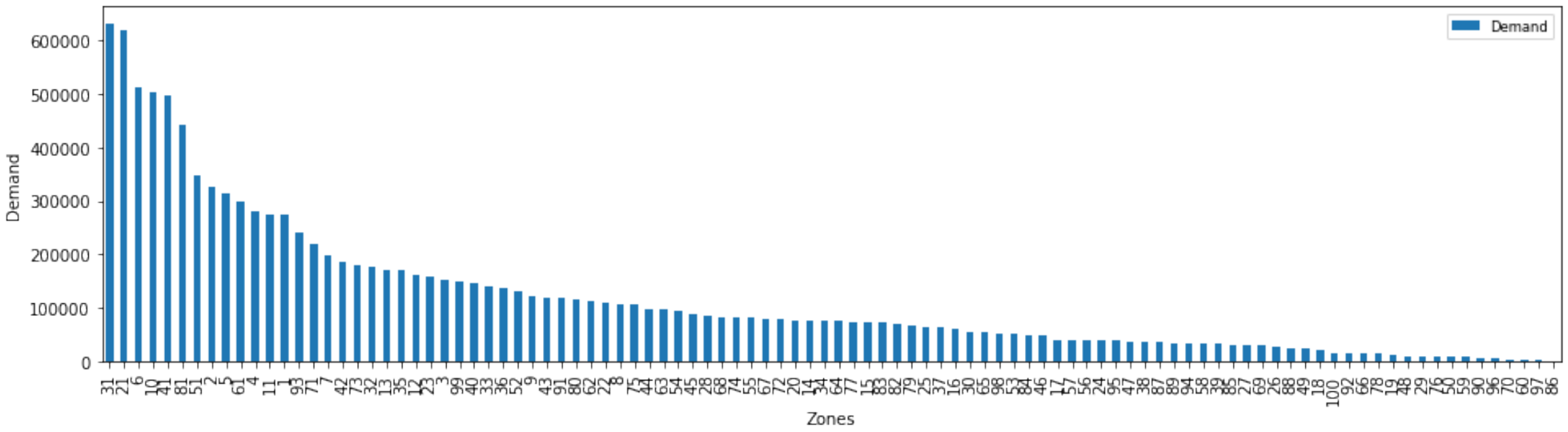

(a) Distribution of demand across zones of Chengdu

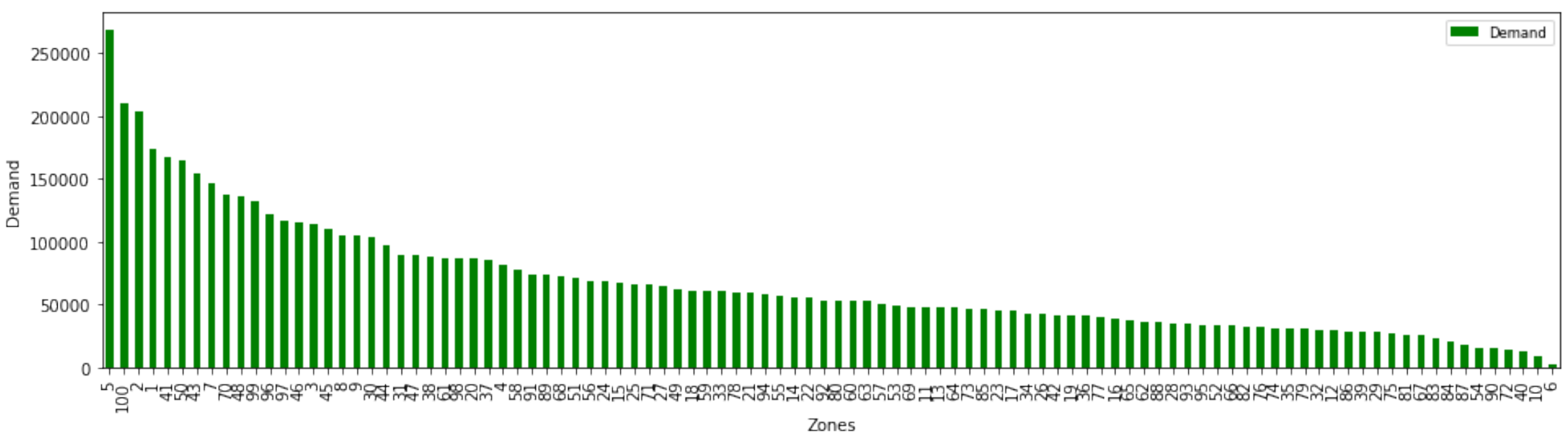

(b) Distribution of demand gap across zones of Xian

Fig. 8. Distribution of demand in Chengdu and Xian

### *5.2. Model Evaluations*

The performance of the proposed GESME-Net is compared against a set of benchmark algorithms. Furthermore, an ablation analysis of GESME-Net is conducted to justify increased model complexity. For a fair comparison, all models utilized the same lookback window, i.e., up to sixth previous time-slot for forecasting original demand and supply-demand gap in the Beijing dataset and up to nineth previous time-slot for forecasting demand in the Chengdu and Xian datasets. The settings of GESME-Net are decided by tuning of the hyperparameters. The finalized settings of the hyperparameters are presented in **Table 1.**

**Table 3.** Hyperparameter settings of GESME-Net

| | **Hyperparameter settings** |
|---|---|
| **(a)** Task Adaptation Layer | Weight initialization: uniform; activation: linear |
| **(b)** Gated Convolutional Mixture of Experts | Layers: 2 layers; experts per layer: 2; filters: 25 in 1$^{st}$ layer; 50 in 2$^{nd}$ layer; filter length: 7-9<br>Weight initialization: uniform |
| **(c)** Gated Recurrent Mixture of Experts | Layers: 2 layers; experts per layer: 2; hidden units: 4 units per layer<br>Weight initialization: uniform |
| **(d)** Gated Convolutional Recurrent Mixture of Experts | Layers: 2 layers; experts per layer: 2; filters: 50 in 1$^{st}$ layer; 100 in 2$^{nd}$ layer; filter length: 5<br>Weight initialization: uniform |
| **(d)** Fully Connected Layer | Weight initialization: uniform; activation: ReLU |
| **(e)** Model Training | Optimizer: Adam (Kingma and Ba, 2015); learning rate: 0.001; batch size: 32<br>Regularization: L1 = 0.001, L2 = 0.001; early stopping patience = 50-100 epochs |

In order to assess the performance of GESME-Net, several machine learning models are considered, which are extensively tuned by utilizing automated machine learning frameworks. They are as follows:

(1) Gradient Boosting Machine (GBM): The GBM (Friedman, 2001) is an ensemble method that is developed from several additive regression trees through the utilization of the gradient descent technique.
(2) Extreme Gradient Boosting (XGBoost): The XGBoost (Chen and Guestrin, 2016), a scalable and more regularized version of GBM, is a popular algorithm utilized in winning many machine learning competitions.
(3) Random Forest (RF): The RF (Liaw and Wiener, 2002) is a bagging ensemble method that utilizes several weak learner regression trees.
(4) Extremely Randomized Trees (XRT): The XRT (Geurts et al., 2006) is similar to the RF except that it extracts randomly generated thresholds for the features.
(5) Generalized Linear Model (GLM): The GLM (Nelder and Wedderburn, 1972) is a generalization of the linear regression that can allow any exponential distributions in the errors.
(6) Artificial Neural Network (ANN): The ANN (Rumelhart et al., 1986) is composed of a neural network with several hidden layers to learn complex patterns from the input features.

In addition to the abovementioned machine learning models, a number of deep learning models are considered as benchmarks. The following configurations are tested:

(1) Spatio-temporal Mixture Network (SM-Net): The SM-Net is a special case of GESME-Net where a single network without subnetworks is utilized without any gating network, specialized to predict only one spatio-temporal task at a time.
(2) Shared-bottom Spatio-temporal Mixture Network (SBSM-Net): The SBSM-Net is similar to SM-Net; however, contains a single network with shared parameters except the tower layer for multi-task learning.
(3) Shared-gated Ensemble of Spatio-temporal Mixture of Experts Network (SESME-Net): The SESME-Net is a variation of GESME-Net, utilizing shared-gating instead of multi-gating for multi-task learning.
(4) Graph Convolutional Network coupled Long Short-Term Memory (GCN-LSTM): The GCN-LSTM (Song et al., 2022) is a graph-based spatio-temporal deep learning model that utilizes non-Euclidean topology using the GCN to extract the spatial features, which are fed into the LSTM model to learn the temporal features.

The performances of the models utilized in this paper are evaluated with three metrics: mean absolute error (MAE), root mean squared error (RMSE), and symmetric mean absolute percentage error (sMAPE), which can be computed by using Eq. (19)-(21):

$$MAE = \frac{1}{n}\sum_{i=1}^{n}|\boldsymbol{O}_i - \boldsymbol{A}_i| \tag{19}$$

$$RMSE = \sqrt{\frac{1}{n}\sum_{i=1}^{n}(\boldsymbol{O}_i - \boldsymbol{A}_i)^2} \tag{20}$$

$$sMAPE = \frac{1}{n}\sum_{i=1}^{n}\frac{|\boldsymbol{O}_i - \boldsymbol{A}_i|}{|\boldsymbol{O}_i| + |\boldsymbol{A}_i| + 1} \tag{21}$$

where $\boldsymbol{O}_i$ and $\boldsymbol{A}_i$ are the predicted vector and ground truth vector, respectively, at time-slot $i$ in the test set with size $n$ time-slots. Since sMAPE produces inaccurate statistics when zero value is encountered in the prediction or ground truth, therefore, a modified sMAPE (Moreira-Matias et al., 2013) is utilized.

Our proposed architecture is trained on a server with 4 Core (hyper-threaded) Xeon processor (2.30 GHz), 25 GB RAM, and a Tesla P-100 GPU. The GESME-Net and its variations are written in Python 3 using Keras (Chollet, 2015) with Tensorflow (Abadi et al., 2015) backend. All the machine learning algorithms are implemented in H2O AutoML (H2O.ai, 2017).

**Table 4.** Performance of GESME-Net and benchmark models for forecasting original demand and supply-demand gap in Beijing (Scenario-1)

| | **Metrics (Original demand)** | | | | **Metrics (Supply-demand gap)** | | | |
|---|---|---|---|---|---|---|---|---|
| **Model** | **MAE** | **RMSE** | **sMAPE** | **Time (s)** | **MAE** | **RMSE** | **sMAPE** | **Time (s)** |
| GESME-Net | 6.41 | 16.83 | 0.1927 | 1198 | 3.42 | 14.91 | 0.2341 | 1198 |

| SESME-Net | 6.43 | 16.95 | 0.1959 | 1185 | 3.51 | 15.11 | 0.2401 | 1185 |
|---|---|---|---|---|---|---|---|---|
| SBSM-Net | 6.58 | 17.20 | 0.1982 | 697 | 3.60 | 15.63 | 0.2577 | 697 |
| SM-Net | 6.68 | 17.00 | 0.1903 | 1094 | 3.52 | 15.49 | 0.2641 | 1936 |
| GCN-LSTM | 6.43 | 16.93 | 0.1953 | 883 | 3.49 | 15.06 | 0.2392 | 957 |
| XGBoost | 6.88 | 19.07 | 0.2009 | 3.68 | 3.70 | 16.74 | 0.2952 | 11.89 |
| GBM | 7.00 | 18.91 | 0.2186 | 3.99 | 3.77 | 16.53 | 0.3053 | 3.43 |
| XRT | 7.04 | 19.43 | 0.1994 | 78.40 | 3.91 | 17.07 | 0.3045 | 31.03 |
| RF | 7.23 | 22.18 | 0.2006 | 79.45 | 3.90 | 16.60 | 0.3044 | 37.32 |
| GLM | 7.65 | 19.78 | 0.2537 | 0.44 | 4.61 | 16.80 | 0.3998 | 1.01 |
| ANN | 14.48 | 25.57 | 0.4250 | 33.21 | 5.34 | 19.09 | 0.4671 | 114.74 |

The performances of the GESME-Net and the benchmark models are reported in **Table 2** and **Table 3**. The GESME-Net performs marginally better than the deep learning benchmarks for each individual task in both scenarios. Such small improvement is not unexpected since all deep learning benchmarks are variations of the GESME-Net. However, the GESME-Net has around 10 % and 8 % lower RMSE than the best machine learning benchmark GBM for forecasting scenario-1 and scenario-2, respectively. Furthermore, the GESME-Net improves the MAE by at least 6 % than the machine learning benchmarks for both scenarios, with a maximum of at least 12 % in the demand forecasting for Xian. The sMAPE improvement of the GESME-Net than the best machine learning benchmark GBM is found to be around 2-6 %. It is noteworthy to mention that the total training times of the multi-task learning models are equally divided among the forecasting tasks to facilitate comparison with the single-task learning models.

**Table 5.** Performance of GESME-Net and benchmark models for forecasting demand in Chengdu and Xian (Scenario-2)

| | **Metrics (Demand-Chengdu)** | | | | **Metrics (Demand-Xian)** | | | |
|---|---|---|---|---|---|---|---|---|
| **Model** | **MAE** | **RMSE** | **sMAPE** | **Time (s)** | **MAE** | **RMSE** | **sMAPE** | **Time (s)** |
| GESME-Net | 3.72 | 5.90 | 0.1574 | 4108 | 2.78 | 4.04 | 0.1632 | 4108 |
| SESME-Net | 3.75 | 6.02 | 0.1574 | 2349 | 2.88 | 4.11 | 0.1698 | 2349 |
| SBSM-Net | 3.82 | 6.18 | 0.1553 | 1523 | 2.84 | 4.11 | 0.1702 | 1523 |
| SM-Net | 3.74 | 5.93 | 0.1549 | 3079 | 2.87 | 4.20 | 0.1679 | 3263 |
| GCN-LSTM | 3.75 | 5.98 | 0.1557 | 1342 | 2.85 | 4.10 | 0.1685 | 1567 |
| XGBoost | 4.15 | 6.53 | 0.1831 | 27.68 | 3.34 | 4.58 | 0.2120 | 26.13 |
| GBM | 4.06 | 6.40 | 0.1805 | 20.80 | 3.18 | 4.44 | 0.2023 | 11.43 |
| XRT | 4.33 | 6.76 | 0.1889 | 133.89 | 3.45 | 4.79 | 0.2127 | 126.96 |
| RF | 4.51 | 6.92 | 0.2000 | 114.41 | 3.25 | 4.68 | 0.1942 | 113.02 |
| GLM | 4.34 | 7.04 | 0.1843 | 1.68 | 3.17 | 4.58 | 0.1922 | 1.76 |
| ANN | 4.32 | 6.83 | 0.2009 | 22.22 | 3.06 | 4.45 | 0.1808 | 18.53 |

Model ablation of GESME-Net is also conducted by removing the model components one at a time. The results of ablation analysis are presented in **Table 4** and **Table 5**. For both scenarios, removal of model components deteriorates the MAE and RMSE, which indicates that each of the model components are essential for achieving best performance of the GESME-Net. The highest deterioration for scenario-1 is seen due to removal of Conv-ME, about 7.5 % increase in the RMSE in the original demand forecasting task and around 10.5 % increase in RMSE in the supply-demand forecasting task. However, the highest deterioration for scenario-2 is found due to removal of GRU-ME, around 4 % increase in the RMSE is seen for both Chengdu and Xian.

**Table 6.** Ablation analysis of GESME-Net in scenario-1

| | **Metrics (Original demand)** | | | **Metrics (Supply-demand gap)** | | | |
|---|---|---|---|---|---|---|---|
| **Removed** | **MAE** | **RMSE** | **sMAPE** | **MAE** | **RMSE** | **sMAPE** | **Time (s)** |
| Weighting | 6.51 | 17.04 | 0.1805 | 3.50 | 15.33 | 0.2464 | 3095 |
| ConvRNN-ME | 6.49 | 17.05 | 0.1990 | 3.46 | 15.32 | 0.2507 | 2770 |
| Conv-ME | 6.89 | 18.09 | 0.2150 | 3.84 | 16.48 | 0.2549 | 5097 |
| ZoneDist(GRU)-ME | 6.69 | 16.96 | 0.2180 | 3.49 | 15.12 | 0.2516 | 1901 |
| GRU-ME | 6.62 | 16.97 | 0.1943 | 3.52 | 14.93 | 0.2543 | 1471 |

**Table 7.** Ablation analysis of GESME-Net in scenario-2

| | **Metrics (Demand-Chengdu)** | | | **Metrics (Demand-Xian)** | | | |
|---|---|---|---|---|---|---|---|
| **Removed** | **MAE** | **RMSE** | **sMAPE** | **MAE** | **RMSE** | **sMAPE** | **Time (s)** |
| Weighting | 3.78 | 6.02 | 0.1583 | 2.85 | 4.10 | 0.1746 | 3146 |
| ConvRNN-ME | 3.71 | 5.96 | 0.1520 | 2.86 | 4.12 | 0.1690 | 4293 |
| Conv-ME | 3.78 | 6.09 | 0.1560 | 2.91 | 4.18 | 0.1717 | 3990 |
| ZoneDist(GRU)-ME | 3.72 | 5.92 | 0.1548 | 2.85 | 4.13 | 0.1719 | 3686 |

| GRU-ME | 3.78 | 6.13 | 0.1553 | 2.97 | 4.21 | 0.1837 | 3457 |
|---|---|---|---|---|---|---|---|

### *5.3. Model Interpretations*

In this paper, the weights utilized by the task adaptation layer in the GESME-Net can be utilized to interpret the contribution of the input features in the forecasting model. In order to separately explain the contribution of the features temporally and spatially, the outputs of the task adaptation layer are averaged spatially and temporally, respectively.

**Fig. 9 (a)** and **Fig. 9 (b)** presents the spatially averaged feature weights across the time-slots for the spatio-temporal features and the weather features for scenario-1 and scenario-2 respectively. For scenario-1, the impotance of the features are almost same in all the time-slots. However, within a time-slot, the spatio-temporal features have more effect than the weather features. For scenario-2, it is interesting to see that the historical values of the spatio-temporal variables, i.e., demand, speed and accessibility, show similar temporal importance patterns in both the cities, are gradually decaying in the previous time-slots and then increasing again. However, this is not the case for the weather variables, the importance of these varibles gradually diminish in the previous time-slots and visibility even have negative association with prediction in the fourth, fifth and sixth previous time-slots. This also indicates that weather variables have relatively less recurrent relationships with the prediction in multi-task learning across cities. Intuitively, weather variables can be city specific and spatio-temporal variables are can have commonalities across cities, which is well distinguished by our GESME-Net models.

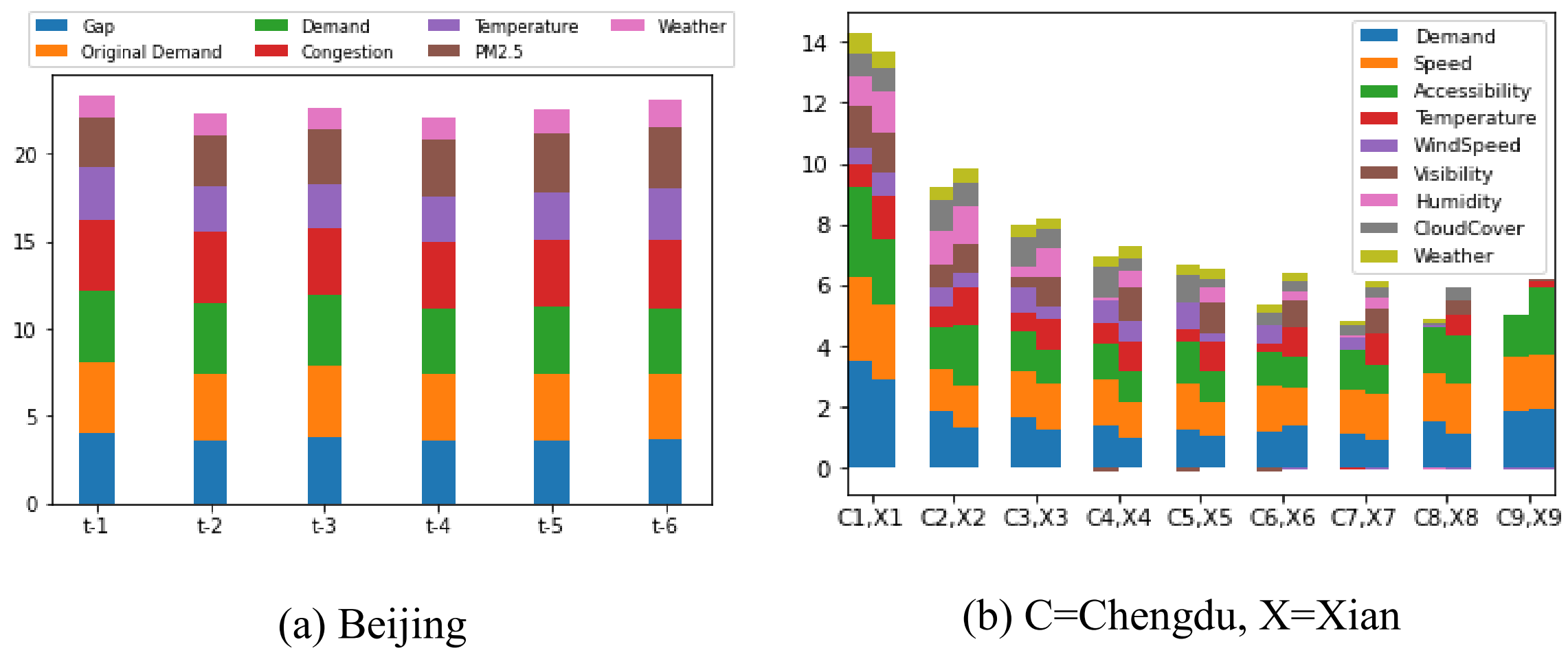

(a) Beijing

(b) C=Chengdu, X=Xian

Fig. 9. Spatially averaged feature importance across previous time-slots

The temporally averaged weights of the spatio-temporal and the context features across the zones are presented in **Fig. 10 (a)** for scenario-1 and in **Fig. 10 (b)** and **Fig. 10 (c)** for scenario-2. It is not unexpected that importance of the spatio-temporal features and the context features are not uniformly weighted across the zones. For both forecasting scenarios, relatively more non-uniform effects are seen for the context features. For scenario-1, in general, it is seen that the features of the higher demand and higher supply-demand gap zones are more important to the GESME-Net for prediction than that of the lower demand and lower supply-demand gap zones. For scenario-2, the spatio-temporal and context variables of the south-east zones in both cities being the most important for prediction, which are usually found to be higher demand zones as evident from **Fig. 8**. The temporal context variables, i.e., peak, off-peak, and sleep, are relatively more emphasized in these zones by our GESME-Net models, which is interesting since we generally expect that time-of-day characteristics are relatively more distinctive in the higher demand zones. Furthermore, GESME-Net also identified POI as an important spatial indicator for prediction in most of the zones and this is more evident in Xian than Chengdu.

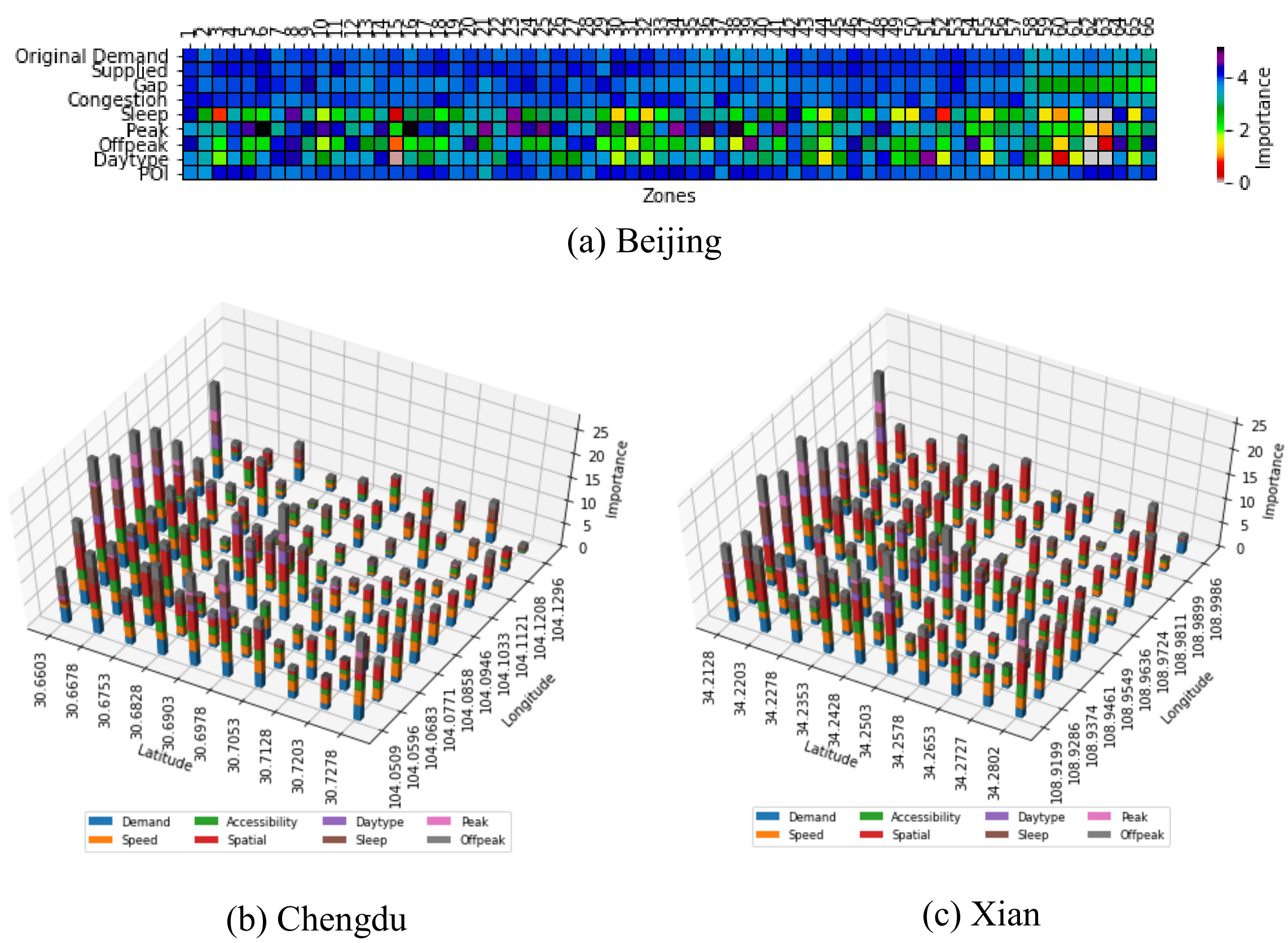

Fig. 10. Temporally averaged feature importance across zones

### *5.4. Sensitivity Analysis*

A sensitivity analysis of GESME-Net is conducted in terms of four hyperparameters: number of layers, number of filters in ConvRNN-ME and Conv-ME, filter size of ConvRNN-ME and Conv-ME, and the number of hidden units in GRU-ME and ZoneDist (GRU)-ME. In order to get the best performance for every hyperparameter configuration, all the experiments on hyperparameter tuning are conducted with early stopping.

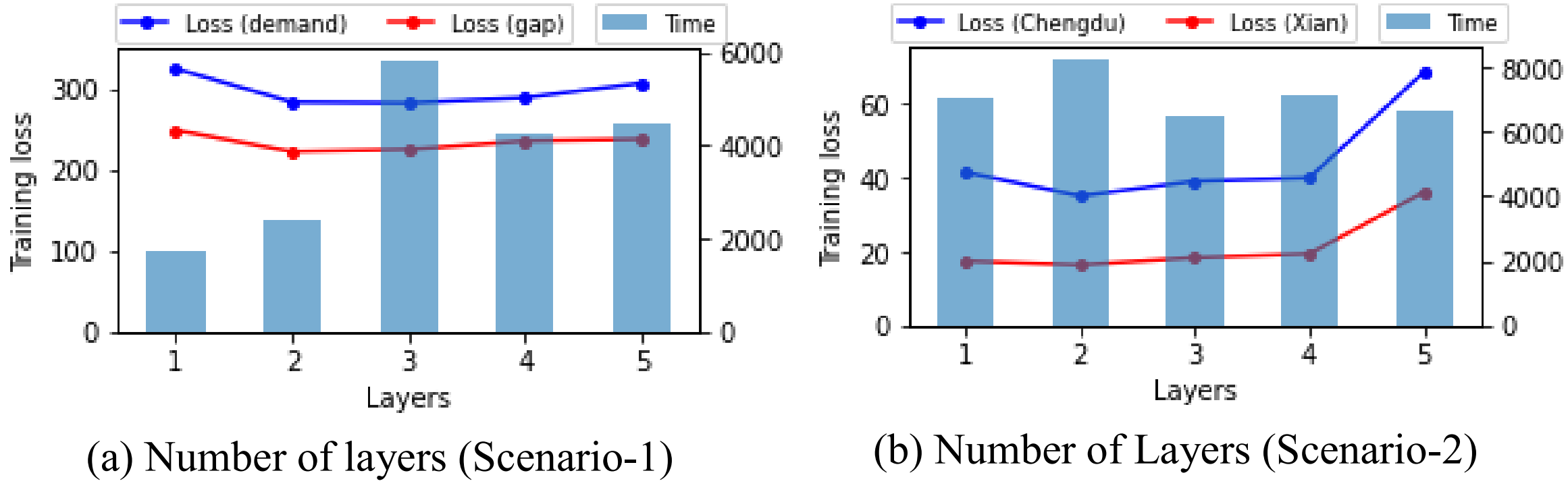

Fig. 11. Sensitivity analysis for number of layers

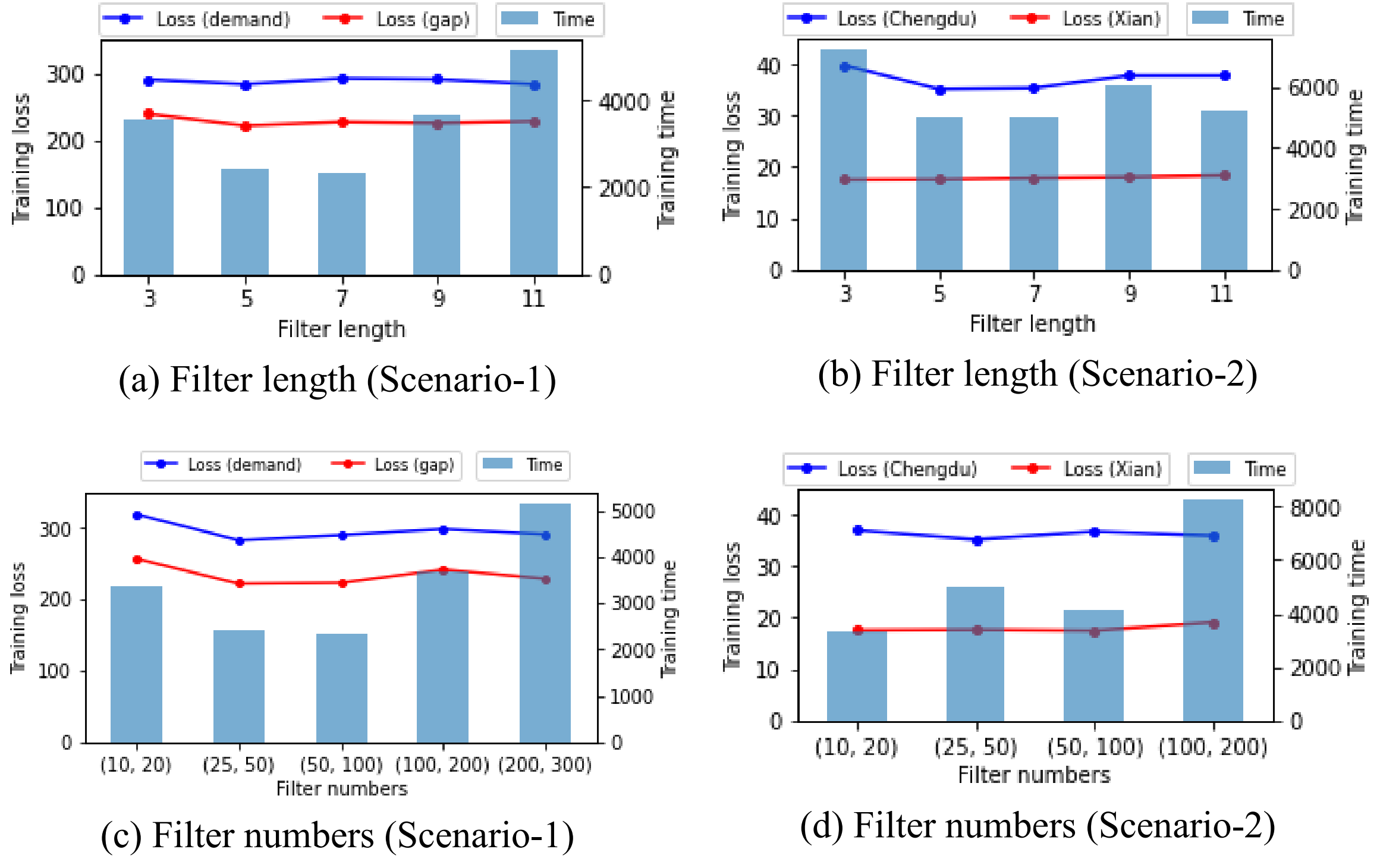

Fig. 12. Sensitivity analysis for ConvRNN-ME

The variation of the training loss (i.e., mean squared error) of GESME-Net with respect to the number of layers for scenario-1 and scenario-2 are shown in **Fig. 11 (a)** and **Fig. 11 (b) respectively**. Losses for both scenarios reach the lowest for 2 layers and then rise again with an increased number of layers. This result is not unintuitive since it is theoretically proven that neural networks with 2 hidden layers can approximate any continuous function (Heaton, 2005). Furthermore, the stability of GESME-Net for higher number of layers demonstrate its suitability for building deeper models.

The performance of the deep learning models with respect to the filter length and the number of filters in the first and second layer of the convolution utilized in ConvRNN-ME are shown in **Fig. 12**. It is evident that the combined loss for the tasks in both scenarios are lowest for filter length 5. Moreover, combined training losses of our proposed architecture for both scenarios gradually decrease and achieves the minimum for a combination of 25 filters in the first layer and 50 filters in the second layer. However, variable effects of filter numbers and filter length on the training losses are seen for convolution utilized in Conv-ME of scenario-1 and scenario-2, as shown in **Fig. 13**. The lowest combined loss is found for filter length 7 and filter combination (50,100) in scenario-1, whereas it is filter length 9 and filter combination (200,300) in scenario-2. It is evident that Conv-ME in our proposed architecture plays a major role in learning task relationships in spatio-temporal multi-task learning.

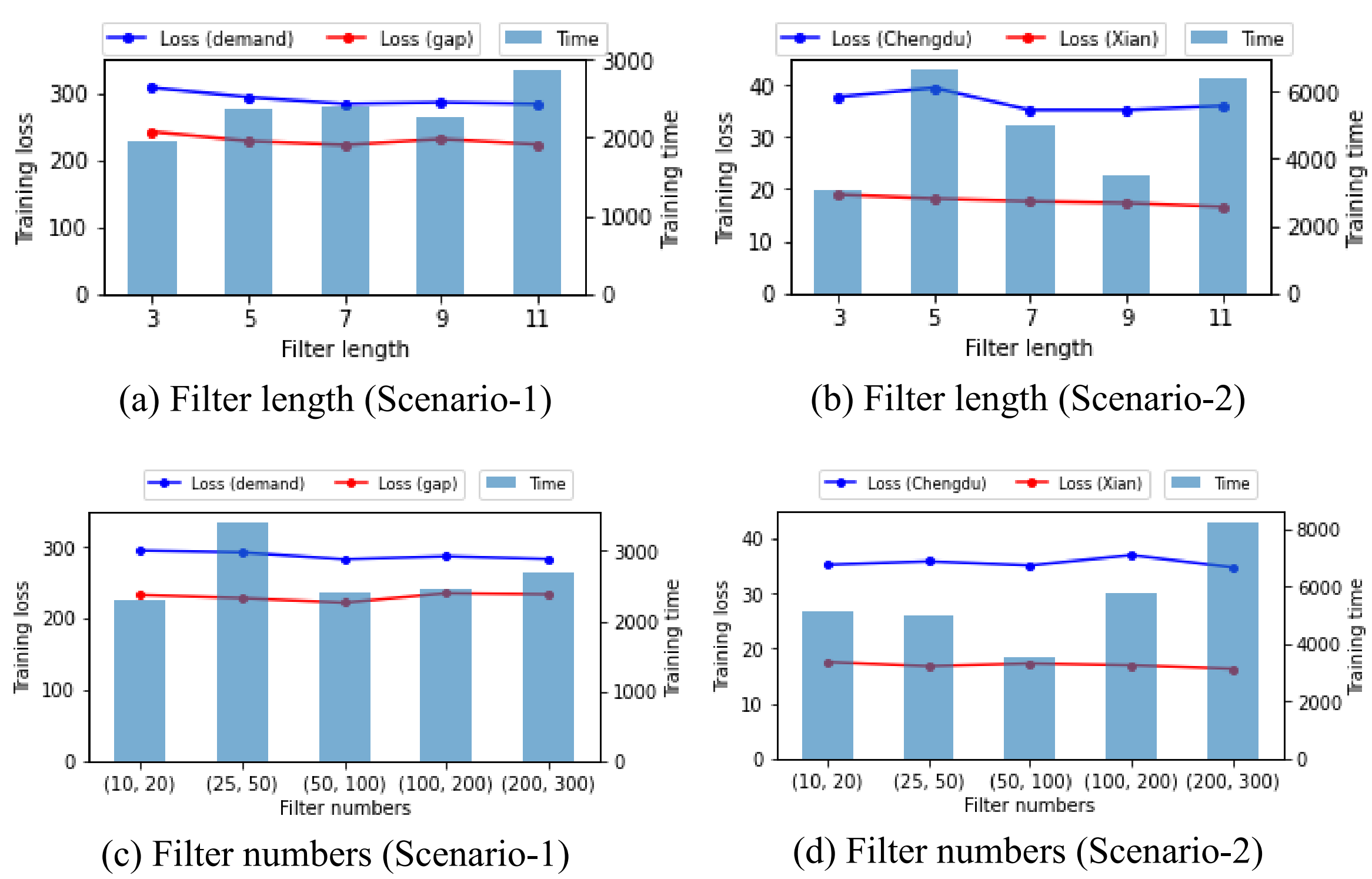

(a) Filter length (Scenario-1) (b) Filter length (Scenario-2)

(c) Filter numbers (Scenario-1) (d) Filter numbers (Scenario-2)

Fig. 13. Sensitivity analysis for Conv-ME

The sensitivity analysis for the variation of hidden units in GRU-ME and ZoneDist(GRU)-ME are shown in Fig. 14. For both scenarios, a common best performance achieved by 4 hidden units. Comparing with other hyperparameters, variations of the hidden units in GRU-ME show less fluctuations in training loss. However, abrupt fluctuations are seen for higher hidden units of ZoneDist(GRU)-ME due to overfitting.

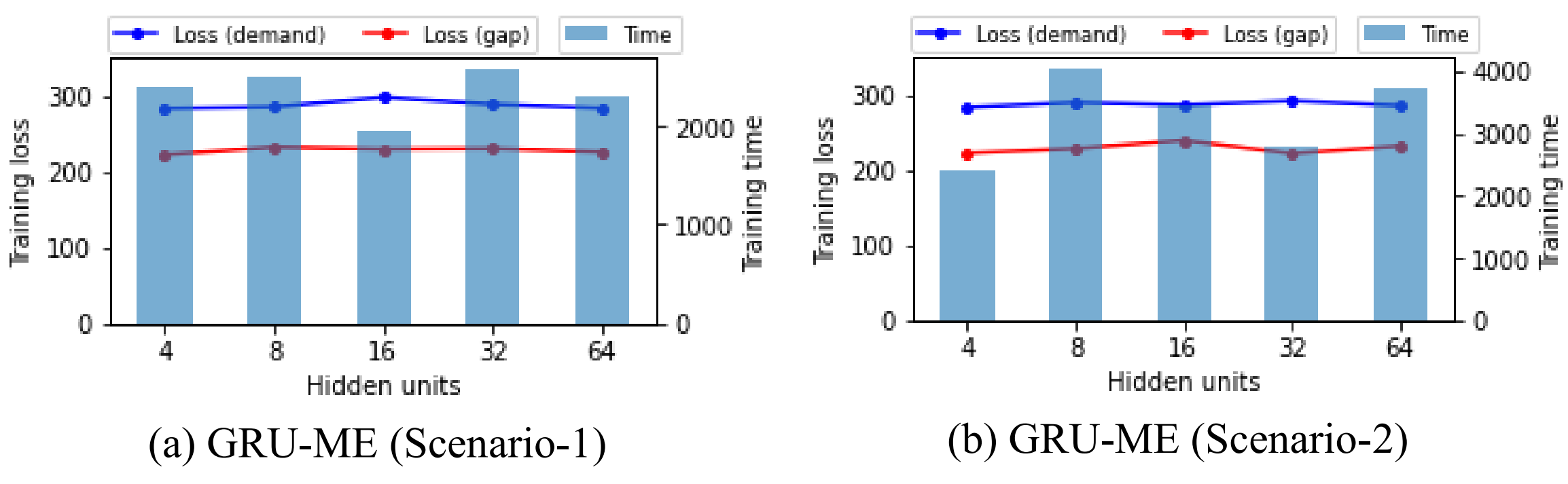

(a) GRU-ME (Scenario-1) (b) GRU-ME (Scenario-2)

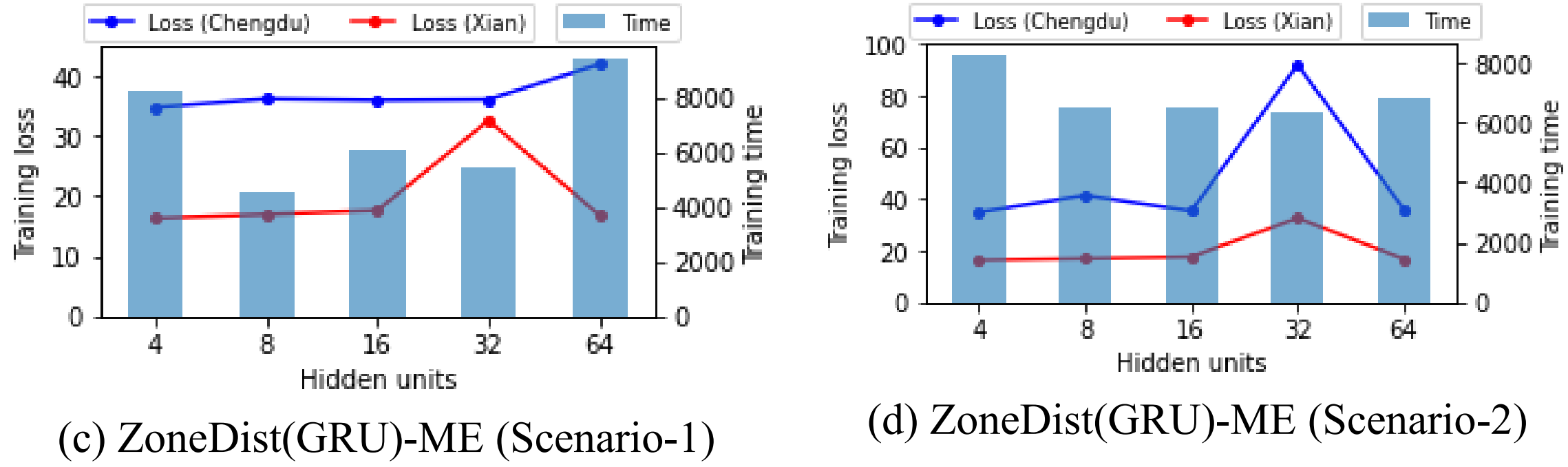

(c) ZoneDist(GRU)-ME (Scenario-1)

(d) ZoneDist(GRU)-ME (Scenario-2)

Fig. 14. Sensitivity analysis for GRU-ME and ZoneDist(GRU)-ME

## 6. Conclusions

In this paper, a spatio-temporal multi-task learning architecture, GESME-Net, is proposed for simultaneously forecasting multiple spatio-temporal tasks in a city as well as across cities. The proposed architecture integrates task adaptation layer with gated ensemble of spatio-temporal mixture of experts, i.e., Conv-ME, GRU-ME, and ConvRNN-ME, to model task relationships in multi-task learning as well as spatio-temporal dependencies. The weights learned from the task adaptation layer assists in learning a common representation in multi-task learning, which can be further utilized in explaining the contribution of the features in the forecasting model. The proposed GESME-Net is compared against several benchmark models including task-specific spatio-temporal deep learning models, spatio-temporal multi-task learning models, and popular machine learning algorithms such as XGBoost, GBM, RF, XRT, GLM, and ANN. The models are tested in two multi-task learning scenarios with real world data from Didi Chuxing, which shows the superiority of GESME-Net in terms of MAE, RMSE, and sMAPE for simultaneously forecasting different spatio-temporal tasks in a city as well as same spatio-temporal task across different cities. An ablation analysis of the GESME-Net is conducted, which shows that each of the model components are crucial for getting best performance from the GESME-Net. Finally, interpretations of the spatio-temporal multi-task learning models are provided based on the outputs of the task adaptation layers. Our analysis provides several important insights into the dynamics of ride-hailing:

- By utilizing a common set of spatio-temporal features for both demand and supply-demand gap predictions within a unified architecture, there is potential for higher accuracy. This occurs because the model can learn from the interdependencies and shared variances between these two elements, which might be overlooked when models are trained separately. Such an approach allows for a more comprehensive understanding of the factors that influence both demand and supply simultaneously.

- A unified model can facilitate learning from data across different cities, capturing broader patterns and anomalies that might not be apparent when models are city-specific. This can be particularly beneficial for ride-hailing companies operating in multiple regions, as the model can adapt to new or changing conditions across different urban environments more effectively.

- Our findings from the model interpretations reveal that not only do the spatio-temporal features such as demand, speed, and accessibility vary in their importance over time, but their impact is also distinct across different city zones. This suggests that effective strategies for managing ride-hailing services must consider both the temporal dynamics of demand and the unique characteristics of different urban areas.

- The model interpretations indicate that weather variables generally have a lesser and decreasing impact on the prediction outcomes as compared to spatio-temporal variables. Interestingly, in some cases, weather conditions like visibility show a negative association with predictions in earlier time-slots. This insight could suggest that ride-hailing demand is more robust against certain weather conditions than expected, or that users' sensitivity to weather variations may decrease with time.

- The model interpretations also points out that in high-demand zones, time-of-day characteristics (peak, off-peak, sleep times) and points of interest (POIs) are particularly significant predictors. This implies that interventions aimed at smoothing demand spikes or enhancing service availability during peak times could be especially effective in these areas.

The proposed architecture demonstrates the viability of utilizing spatio-temporal multi-task learning architecture for jointly forecasting demand and supply-demand gap in a ride-hailing system. Furthermore, model interpretation from the task adaptation layer can assist in learning joint representation in spatio-temporal multi-task learning. Nevertheless, our has some limitations. In this study, our analysis was constrained to the utilization of only two external variables: weather conditions and points of interest (POI). This limitation was primarily due to restricted availability of external data for the specific areas under investigation. Detailed information for some of the features such as weather categories, traffic congestion levels, and POI are unavailable in the Beijing datasets due to confidentiality and privacy issues that restricted us from utilizing more features. Further improvement can be achieved by incorporating a wider range of external variables, such as socio-economic economic indicators, special events, and public transport information. In addition to limited external features, taxi trajectory data from a limited area of the Chengdu and Xian city are available, which limited us from large scale testing. Further investigation can be done when city-wide ride-hailing trajectory data becomes available. The proposed architecture can be tested against a large number of spatio-temporal tasks in the future when data from several cities are available.

## Acknowledgments

Comments from anonymous reviewers and editors have led to significant improvements to this paper. The authors are grateful for this assistance. The authors are grateful to Didi Chuxing for the publicly available datasets (https://research.xiaojukeji.com/) and the open datasets datasets under the GAIA Open Dataset Initiative (https://gaia.didichuxing.com).